\documentclass[10pt,twocolumn,letterpaper]{article}

\usepackage[pagenumbers]{cvpr} 

\usepackage{graphicx}
\usepackage{amsmath}
\usepackage{amssymb}
\usepackage{booktabs}
\usepackage{multirow}
\usepackage{rotating}
\usepackage[table,xcdraw,x11names]{xcolor}
\usepackage[normalem]{ulem}
\useunder{\uline}{\ul}{}
\usepackage{epigraph} 
\usepackage{dsfont}
\usepackage{bm}
\usepackage{enumitem} 
\usepackage[accsupp]{axessibility}

\usepackage[pagebackref,breaklinks,colorlinks]{hyperref}

\usepackage[capitalize]{cleveref}
\crefname{section}{Sec.}{Secs.}
\Crefname{section}{Section}{Sections}
\Crefname{table}{Table}{Tables}
\crefname{table}{Tab.}{Tabs.}

\newcommand{\pose}{\mathbf{h}}
\newcommand{\crd}{\mathbf{y}}
\newcommand{\eye}{\mathbf{e}}
\newcommand{\pos}{\mathbf{x}}
\newcommand{\patch}{\mathbf{p}}
\newcommand{\proj}{\boldsymbol{\pi}}
\newcommand{\param}{\mathbf{w}}
\newcommand{\feat}{\mathbf{f}}

\newcommand{\greencell}{\cellcolor{Green2!25}}
\newcommand{\orangecell}{\cellcolor{Orange1!40}}
\newcommand{\redcell}{\cellcolor{IndianRed1!40}}

\DeclareMathOperator*{\argmin}{\arg\!\min}

\def\cvprPaperID{****} 
\def\confName{CVPR}
\def\confYear{2023}

\begin{document}

\title{Accelerated Coordinate Encoding: \\ Learning to Relocalize in Minutes using RGB and Poses}

\author{Eric Brachmann\\
Niantic
\and
Tommaso Cavallari\\
Niantic
\and
Victor Adrian Prisacariu\\
Niantic, University of Oxford
}

\twocolumn[{%
\renewcommand\twocolumn[1][]{#1}%
\maketitle
\begin{center}
    \centering
    \captionsetup{type=figure}
    \includegraphics[width=1.0\linewidth]{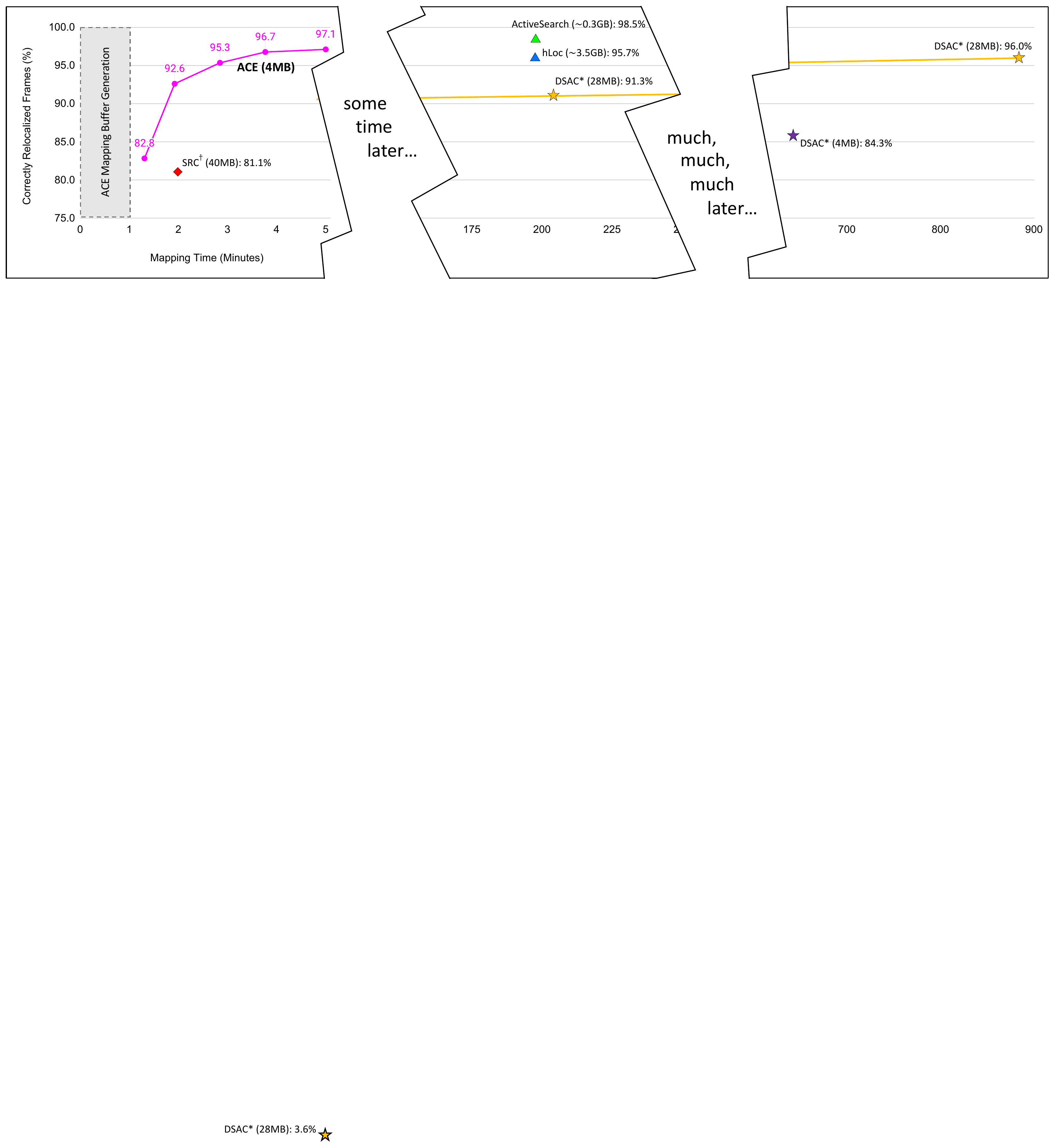}
    \vspace{-15.0cm}
    \captionof{figure}{\textbf{Mapping Time vs.~Relocalization Rate.} We show the mapping time of multiple state-of-the-art relocalizers on a standard dataset, 7Scenes \cite{shotton2013scene}. We measure accuracy as the percentage of frames with a pose error below 5cm and 5$^\circ$. Our approach, ACE, maps a new environment two orders of magnitude faster than the baseline, DSAC* \cite{Brachmann2021dsacstar}, while being as accurate. We also report the average map size of each approach in brackets. $\dag$ signifies that mapping needs depth, while other methods only need RGB and poses. }
    \label{fig:teaser}
\end{center}%
\vspace{0.5cm}
}]
\maketitle

\begin{abstract}
\vspace{-0.3cm}
Learning-based visual relocalizers exhibit leading pose accuracy, but require hours or days of training.
Since training needs to happen on each new scene again, long training times make learning-based relocalization impractical for most applications, despite its promise of high accuracy.
In this paper we show how such a system can actually achieve the same accuracy in less than 5 minutes.
We start from the obvious: a relocalization network can be split in a scene-agnostic feature backbone, and a scene-specific prediction head.
Less obvious: using an MLP prediction head allows us to optimize across thousands of view points simultaneously in each single training iteration.
This leads to stable and extremely fast convergence.
Furthermore, we substitute effective but slow end-to-end training using a robust pose solver with a curriculum over a reprojection loss. 
Our approach does not require privileged knowledge, such a depth maps or a 3D model, for speedy training.
Overall, our approach is up to 300x faster in mapping than state-of-the-art scene coordinate regression, while keeping accuracy on par. 
Code is available: \url{https://nianticlabs.github.io/ace}
\end{abstract}

\section{Introduction}
\label{sec:intro}

\epigraph{Time is really the only capital that any human being has, and the only thing he can’t afford to lose.}{Thomas Edison}

Time is relative.
Time spent waiting can stretch to infinity. 
Imagine waiting for a visual relocalizer to finally work in a new environment.
It can take hours -- and feel like days -- until the relocalizer has finished its pre-processing of the scene.
Only then can it estimate the camera's position and orientation to support real-time applications like navigation or augmented reality (AR). 

Relocalizers need that extensive pre-processing to build a map of the environment that defines the coordinate space we want to relocalize in.
Visual relocalizers typically build maps from sets of images of the environment, for each of which the camera pose is known.
There are two prevalent families of structure-based relocalizers that meet the high accuracy requirements of applications like AR.

Sparse feature-matching approaches \cite{sattler2012improving, Sattler2017AS, compression2019cvpr, sarlin2019coarse, kapture2020, zhou2022gomatch, Panek2022meshloc} need to build an explicit 3D reconstruction of a scene using structure-from-motion (SfM) software \cite{Snavely2006bundler,wu2011visualsfm,schoenberger2016sfm}.
Even when poses of mapping images are known, the runtime of SfM for scene triangulation varies a lot, and can lie anywhere between 10 minutes and 10 hours depending on how many mapping frames are used.
When mapping succeeds, feature-based relocalizers are fast at query time and accurate \cite{Sattler2017AS, sarlin2019coarse}.
Less refined maps can be built in real time using SLAM, if one is willing to accept the detrimental effect on accuracy \cite{brachmann2021limits}.
In either case, the underlying maps can consume vast amounts of storage, and can reveal private information that was present in the mapping images \cite{Speciale2019privacy, Chelani2021privacyhack}.

On the other hand, scene coordinate regression \cite{shotton2013scene, brachmann2017dsac, Brachmann2018dsacpp, yang2019sanet, li2020hierarchical, Brachmann2021dsacstar, dong2022visual} \emph{learns} an implicit representation of the scene via gradient descent. 
The resulting maps can be as small as 4MB \cite{Brachmann2021dsacstar}, and privacy preserving \cite{zhou2022gomatch}.
But, while scene coordinate regression is on-par with feature-matching in terms of accuracy and relocalization time \cite{brachmann2021limits},
the fact that they map an environment via hours-long \emph{training of a network} makes them unattractive for most applications.
The state-of-the-art scene coordinate regression pipeline, DSAC* \cite{Brachmann2021dsacstar}, requires 15 hours to reach top accuracy on a premium GPU, see Fig.~\ref{fig:teaser}.
We can stop training any time, and see which accuracy we get
but, after 5 minutes mapping time, DSAC* has a relocalization rate in the single digits.
In fact, the corresponding data point for the plot in Fig.~\ref{fig:teaser} can be found at the bottom of the previous page.

The aim of this work is summarized quickly: 
we take a scene coordinate regression-based relocalizer, the slowest approach in terms of mapping time, and make it one of the fastest.
In particular, we present \textit{Accelerated Coordinate Encoding} (ACE), a schema to train scene coordinate regression in 5 minutes to state-of-the-art accuracy.

Speeding up training time normally causes moderate interest in our community, at best.
This is somewhat justified in train-once-deploy-often settings. 
Still, learning-based visual relocalization does not fall within that category, as training needs to happen on each new scene, again.
Therefore, fast training has a range of important implications:

\begin{itemize}[noitemsep]
    \item \textbf{Mapping delay.} We reduce the time between collecting mapping data, and having a top-performing relocalizer for that environment.
    \item \textbf{Cost.} Computation time is expensive. Our approach maps a scene within minutes on a budget GPU. 
    \item \textbf{Energy consumption.} Extensive computation is an environmental burden. We significantly reduce the resource footprint of learning-based relocalization.
    \item \textbf{Reproducibility.} Using ACE to map all scenes of the datasets used in this paper can be done almost five times over on a budget GPU, in the time it takes DSAC* to map a single scene on a premium GPU.
\end{itemize}

\noindent We show that a thoughtful split of a standard scene coordinate regression network allows for more efficient training.
In particular, we regard scene coordinate regression as a mapping from a high-dimensional feature vector to a 3D point in scene space. 
We show that a multi-layer perceptron (MLP) can represent that mapping well, as opposed to convolutional networks normally deployed \cite{Brachmann2018dsacpp, Brachmann2021dsacstar, dong2022visual}. 
Training a scene-specific MLP allows us to optimize over many (oftentimes all available) mapping views at once in each single training iteration.
This leads to very stable gradients that allow us to operate in very aggressive, high-learning rate regimes.
We couple this with a curriculum over a reprojection loss that lets the network \emph{burn in} on reliable scene structures at later stages of training. 
This mimics end-to-end training schemes that involve differentiating through robust pose estimation during training \cite{Brachmann2021dsacstar}, but are much slower than our approach.
\noindent We summarize our \textbf{contributions}:
\begin{itemize}[noitemsep]
    \item \emph{Accelerated Coordinate Encoding} (ACE), a scene coordinate regression system that maps a new scene in 5 minutes. Previous state-of-the-art scene coordinate regression systems require hours of mapping to achieve comparable relocalization accuracy. 
    \item ACE compiles a scene into 4MB worth of network weights. Previous scene coordinate regression systems required 7-times more storage, or had to sacrifice accuracy for scene compression.
    \item Our approach requires only posed RGB images for mapping. Previous fast mapping relocalizers relied on priviledged knowledge like depth maps or a scene mesh for speedy mapping. 
\end{itemize}

\section{Related Work}
\label{sec:relwork}

Visual relocalization requires some representation of the environment we want to relocalize in.
We refer to these representations as ``maps", and the process of creating them as ``mapping".
Our work is predominately concerned with the time needed for mapping, and, secondly, the storage demand of the maps created.

\paragraph{Image Retrieval and Pose Regression.}

Arguably the simplest form of a map is a database of mapping images and their poses. 
Given a query image, we look for the most similar mapping images using image retrieval \cite{torii2015denseVLAD, netvlad, revaud2019apgem}, and approximate the query pose with the top retrieved mapping pose \cite{compression2019cvpr, sattler2019limits}. 
Pose regression uses neural networks to either predict the absolute pose from a query image directly, or predict the relative pose between the query image and the top retrieved mapping image.
All absolute and most relative pose regression methods \cite{kendall2015posenet, Kendall2017GeometricLF, Brahmbhatt2018mapnet, zhou2020essnet, WinkelbauerICRA21, turkoglu2021visual, Shavit2021MStransformer} train scene-specific networks which can take significant time, \eg \cite{Kendall2017GeometricLF} reports multiple hours per scene for PoseNet.
Some relative pose regression works report results with generalist, scene-agnostic networks that do not incur additional mapping time on top of building the retrieval index \cite{WinkelbauerICRA21, turkoglu2021visual}.
Map-free relocalization \cite{arnold2022mapfree} is an extreme variation that couples scene-agnostic relative pose regression with a single reference frame for practically instant relocalization. 
Recently, some authors use neural radiance fields (NeRFs) \cite{mildenhall2020nerf} for camera pose estimation \cite{yen2020inerf,maggio2022locnerf}.
In its early stage, this family of methods has yet to demonstrate its merits against the corpus of existing relocalisers and on standard benchmarks.
Some of the aforementioned approaches have attractive mapping times, \ie require only little scene-specific pre-processing.
But their pose accuracy falls far behind structure-based approaches that we discuss next.

\paragraph{Feature Matching.}

Feature matching-based relocalizers \cite{Sattler2017AS, sarlin2019coarse, compression2019cvpr, kapture2020,Panek2022meshloc} calculate the camera pose from correspondences between the query image and 3D scene space.
They establish correspondences via discrete matching of local feature descriptors.
Thus, they require a 3D point cloud of an environment where each 3D point stores one or multiple feature descriptors for matching.
These point clouds can be created by running SfM software, such as COLMAP \cite{schoenberger2016sfm}.
Even if poses of mapping images are known in advance, \eg from on-device visual odometry \cite{newcombe2011kinectfusion, izadi2011kinectfusion, arkit, arcore}, feature triangulation with SfM can take several hours, depending on the number of mapping frames.
Also, the storage requirements can be significant, mainly due to the need for storing hundreds of thousands of descriptor vectors for matching. 
Strategies exist to alleviate the storage burden, such as storing fewer descriptors per 3D point \cite{Irschara2009sfmfast, sattler2011fast, Sattler2017AS}, compressing descriptors \cite{lynen2020visin,Yang2022squeezer} or removing 3D points \cite{Yang2022squeezer}.
More recently, GoMatch \cite{zhou2022gomatch} and MeshLoc \cite{Panek2022meshloc} removed the need to store descriptors entirely by matching against the scene geometry.
None of the aforementioned strategies reduce the mapping time -- on the converse, often they incur additional post-processing costs for the SfM point clouds.
To reduce mapping time, one could use only a fraction of all mapping images for SfM, or reduce the image resolution.
However, this would likely also affect the pose estimation accuracy.

\paragraph{Scene Coordinate Regression.}
Relocalizers in this family regress 3D coordinates in scene space for a given 2D pixel position in the query image \cite{shotton2013scene}.
Robust optimization over scene-to-image correspondences yields the desired query camera pose. 
To regress correspondences, most works rely on random forests \cite{shotton2013scene, valentin2015cvpr, brachmann2016, cavallari2017fly, Cavallari2019cascade} or, more recently, convolutional neural networks \cite{brachmann2017dsac, Brachmann2018dsacpp, Brachmann2021dsacstar, Brachmann2019ESAC, Cavallari2019network, li2020hierarchical, dong2022visual}. 
Thus, the scene representation is implicit, and the map is encoded in the weights of the neural network.
This has advantages as the implicit map is privacy-preserving \cite{Speciale2019privacy, zhou2022gomatch}:
an explicit scene representation can only be re-generated with images of the environment. 
Also, scene coordinate regression has small storage requirements.
DSAC* \cite{Brachmann2021dsacstar} achieves state-of-the-art accuracy with 28MB networks, and acceptable accuracy with 4MB networks.
Relocalization in large-scale environments can be challenging, but strategies exist that rely on network ensembles \cite{Brachmann2019ESAC}.

The main drawback of scene coordinate regression is its long mapping time, since mapping entails training a neural network for each specific scene.
DSAC++\cite{Brachmann2018dsacpp} reported 6 days of training for a single scene. 
DSAC* reduced the training time to 15 hours -- given a powerful GPU.
This is still one order of magnitude slower than typical feature matching approaches need to reconstruct a scene. 
In our work, we show how few conceptual changes to a scene coordinate regression pipeline result in a speedup of two orders of magnitude. 
Thus, we pave the way for deep scene coordinate regression to be useful in practical applications.

A variety of recipes allow for fast mapping if depth, rendered or measured, is given.
Indeed, the original SCoRF paper \cite{shotton2013scene} reported to train their random forest with \mbox{RGB-D} images under 10 minutes.
Cavallari \etal \cite{Cavallari2019cascade} show how to adapt a pre-trained neural scene representation in real time for a new scene, but their approach requires depth inputs for the adaptation, and for relocalization.
Dong \etal \cite{dong2022visual} use very few mapping frames with depth to achieve a mapping time of 2 minutes. 
The architecture described in \cite{dong2022visual} consists of a scene-agnostic feature backbone, and a scene-specific region classification head -- very similar to our setup. 
However, their prediction head is convolutional, and thus misses the opportunity for highly efficient training as we will show.
SANet \cite{yang2019sanet} is a scene coordinate regression variant that builds on image retrieval.
A scene-agnostic network interpolates the coordinate maps of the top retrieved mapping frames to yield the query scene coordinates. 
None of the aforementioned approaches is applicable when mapping images are RGB only.
Depth channels for mapping can be rendered from a dense scene mesh \cite{Brachmann2018dsacpp}, but mesh creation would increase the mapping time.
Our work is the first to show fast scene coordinate regression mapping from RGB and poses alone. 

\section{Method}
\label{sec:method}

Our goal is to estimate a camera pose $\pose$ given a single RGB image $I$.
We define the camera pose as the rigid body transformation that maps coordinates in camera space $\eye_i$ to coordinates in scene space $\crd_i$, therefore $\crd_i = \pose \eye_i$.
We can estimate the pose from image-to-scene correspondences:
\begin{equation}
\pose = g(\mathcal{C}),~\text{with}~\mathcal{C}=\{(\pos_i, \crd_i)\},
\end{equation}
where $\mathcal{C}$ is the set of correspondences between 2D pixel positions $\pos_i$ and 3D scene coordinates $\crd_i$.
Function $g$ denotes a robust pose solver.
Usually $g$ consists of a PnP minimal solver \cite{gao2003complete} in a RANSAC \cite{fischler1981random} loop, followed by refinement.
Refinement consists of iterative optimization of the reprojection error over all RANSAC inliers using Levenberg–Marquardt \cite{Levenberg44lm, Marquardt63lm}. 
For more details concerning pose solving we refer to \cite{Brachmann2021dsacstar}, as our focus is on correspondence prediction.
To obtain correspondences, we follow the approach of scene coordinate regression \cite{shotton2013scene}.
We learn a function to predict 3D scene points for any 2D image location:
\begin{equation}
\crd_i = f(\patch_i;\param),~\text{with}~\patch_i=\mathcal{P}(\pos_i, I),
\end{equation}
where $f$ is a neural network parameterized by learnable weights $\param$, and $\patch_i$ is an image patch extracted around pixel position $\pos_i$ from image $I$.
Therefore, $f$ implements a mapping from patches to coordinates, $f:\mathds{R}^{C_\text{I}\times H_\text{P} \times W_\text{P}} \rightarrow \mathds{R}^3$.
We have RGB images but usually take grayscale inputs with $C_\text{I}=1$.
Typical patch dimensions are $H_\text{P}=W_\text{P}=81\text{px}$ \cite{Brachmann2018dsacpp, Brachmann2021dsacstar, Brachmann2019ESAC, brachmann2019NGransac}.
For state-of-the-art architectures there is no explicit patch extraction.
A fully convolutional neural network \cite{fcn2015} with limited receptive field slides over the input image to efficiently predict dense outputs while reusing computation of neighbouring pixels.
However, for our subsequent discussion, the explicit patch notation will prove useful.

\begin{figure}[t]
  \centering
   \includegraphics[width=1.0\linewidth]{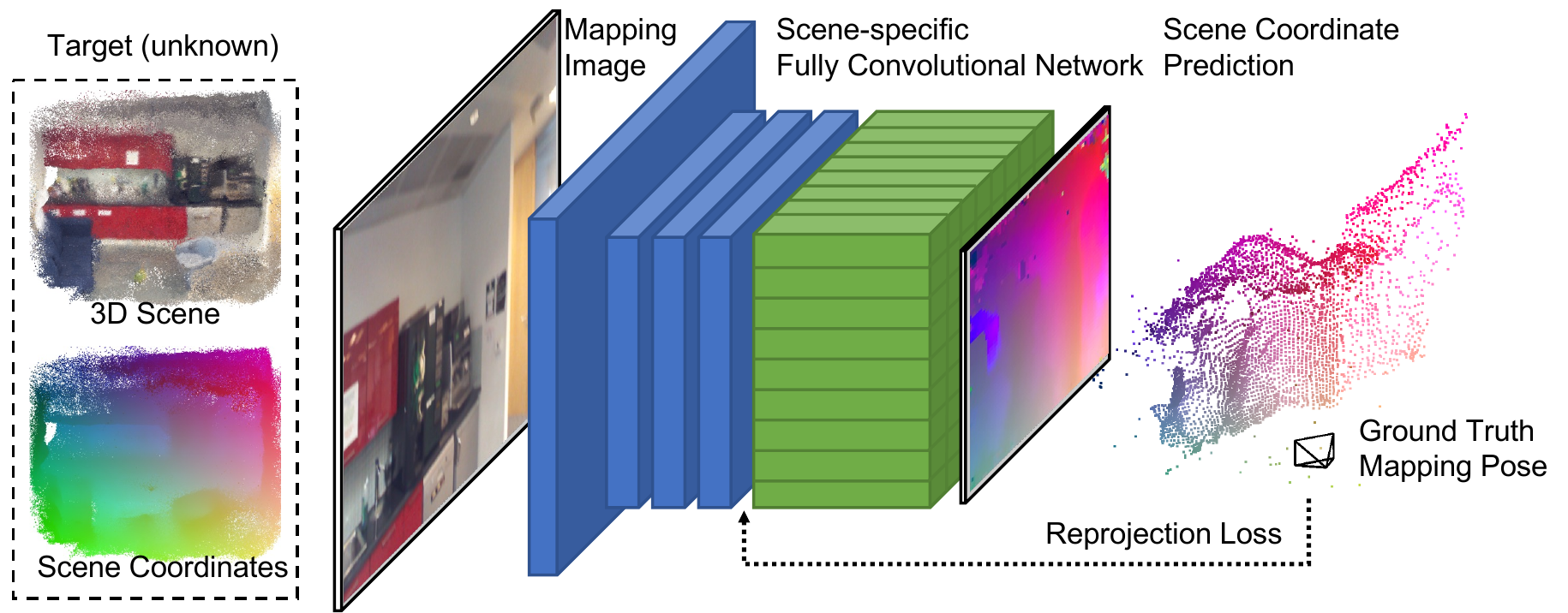}
   \caption{\textbf{Standard Training Loop \cite{Brachmann2021dsacstar}}. Previous works train a coordinate regression network with one mapping image at a time. The network predicts dense scene coordinates, and is supervised with the ground truth camera pose and a reprojection loss. }
   \label{fig:system_dsac}
   \vspace{-1em}
\end{figure}

We learn the function $f$ by optimizing over all mapping images $\mathcal{I}_\text{M}$ with their ground truth poses $\pose_i^*$ as supervision:
\begin{equation}
\label{eq:training_loss}
    \argmin_\param \sum_{I \in \mathcal{I}_\text{M}} \sum_i \ell_{\proj}[\pos_i, \overbrace{f(\patch_i;\param)}^{\crd_i}, \pose_i^*],
\end{equation}
where $\ell_{\proj}$ is a reprojection loss that we discuss in Sec.~\ref{sec:curriculum}.
We optimize Eq.~\ref{eq:training_loss} using minibatch stochastic gradient descent.
The network predicts dense scene coordinates from one mapping image at a time, and all predictions are supervised using the ground truth mapping pose, see  Fig.~\ref{fig:system_dsac}.

\subsection{Efficient Training by Gradient Decorrelation}

With the standard training, we optimize over predictions for thousands of patches in each training iteration -- but they all come from the same image. 
Hence, their loss and their gradients will be highly correlated.
A prediction $\crd_i$ and the prediction for the pixel next to it will be very similar, so will be the pixel loss and its gradient.

Our key idea is to randomize patches over the entire training set, and construct training batches from many different mapping views.
This decorrelates gradients within a batch and leads to a very stable training signal, robustness to high learning rates, and, ultimately, fast convergence.

A naive implementation of this idea would be slow if it resorted to explicit patch extraction \cite{brachmann2017dsac}.
The expressive power of convolutional layers, and their efficient computation using fully convolutional architectures is key for state-of-the-art scene coordinate regression.
Therefore, we propose to split the regression network into a convolutional backbone, and a multi-layer perceptron (MLP) head:
\begin{equation}
f(\patch_i;\param) = f_\text{H}(\feat_i; \param_\text{H}),~\text{with}~\feat_i=f_\text{B}(\patch_i; \param_\text{B}),
\end{equation}
where $f_\text{B}$ is the backbone that predicts a high-dimensional feature $\feat_i$ with dimensionality $C_\feat$, and $f_\text{H}$ is the regression head that predicts scene coordinates:
\begin{equation}
    f_\text{B}: \mathds{R}^{C_\text{I}\times H_\text{P} \times W_\text{P}} \rightarrow \mathds{R}^{C_\feat}~\text{and}~f_\text{H}: \mathds{R}^{C_\feat} \rightarrow \mathds{R}^{3}.
\end{equation}

Similar to \cite{dong2022visual}, we argue that $f_\text{B}$ can be implemented using a scene-agnostic convolutional network - a generic feature extractor.
In addition to \cite{dong2022visual}, we argue that $f_\text{H}$ can be implemented using a MLP instead of another convolutional network.
Fig.~\ref{fig:system_dsac} signifies our network split. 
Convolution layers with $3\times3$ kernels are blue, and $1\times1$ convolutions are green. 
The latter are MLPs with shared weights. 
This standard network design is used in pipelines like DSAC* \cite{Brachmann2021dsacstar}.

\begin{figure*}[t]
  \centering
   \includegraphics[width=1.0\linewidth]{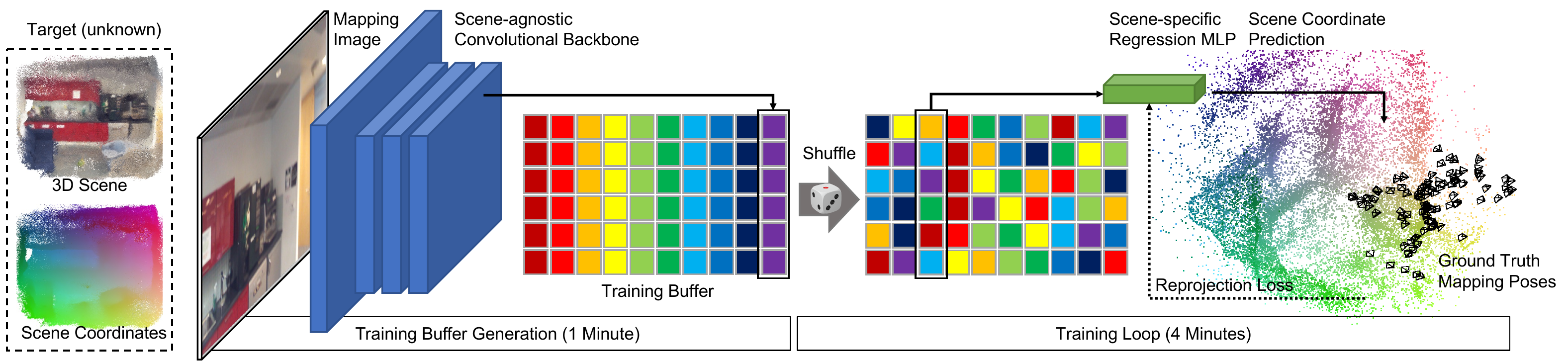}

   \caption{\textbf{ACE Training Loop.} Training consists of two stages: Buffer generation (left) and the main training loop (right). To create a training buffer, we pass mapping images through a scene-agnostic backbone that extracts high-dimensional feature vectors. Each colored box in the buffer represents one such feature, and features with the same color came from the same mapping image. In the main loop, we train a scene-specific MLP that predicts scene coordinates from backbone features. We assemble training batches from random features and their associated mapping poses. Thus, we supervise the scene-specific MLP with many, diverse mapping views in each training iteration.}
   \label{fig:system_ace}
   \vspace{-1em}
\end{figure*}

Note how function $f_\text{H}$ needs no spatial context, \ie differently from the backbone, $f_\text{H}$ does not need access to neighbouring pixels for its computation.
Therefore, we can easily construct training batches for $f_\text{H}$ with random samples across all mapping images.
Specifically, we construct a fixed size training buffer by running the pre-trained backbone $f_\text{B}$ over the mapping images.
This buffer contains millions of features $\feat_i$ with their associated pixels positions $\pos_i$, camera intrinsics $\mathbf{K}_i$ and ground truth mapping poses ${\pose^*_i}$.
We generate this buffer once, in the first minute of training.
Afterwards, we start the main training loop that iterates over the buffer.
At the beginning of each epoch, we shuffle the buffer to mix features (essentially patches) across all mapping data.
In each training step, we construct batches of several thousand features, potentially computing a parameter update over thousands of mapping views at once.
Not only is the gradient computation extremely efficient for our MLP regression head, but the gradients are also decorrelated which allows us to use high learning rates for fast convergence.
Fig.~\ref{fig:system_ace} shows our training procedure.

\subsection{Curriculum Training}
\label{sec:curriculum}

Previous state-of-the-art scene coordinate regression pipelines use a multi-stage training process. 
Firstly, they optimize a pixel-level reprojection loss.
Secondly, they do end-to-end training, where they propagate a pose error back through a differentiable pose solver \cite{brachmann2017dsac, Brachmann2018dsacpp}.
End-to-end training lets the network focus on reliable scene structures while ignoring outlier predictions. 
However, end-to-end training is extremely costly. 
For example, in \cite{Brachmann2021dsacstar}, end-to-end training incurs half of the training time for 10\% of the parameter updates.
To mimic the effects of end-to-end training, we construct a curriculum over a much simpler pixel-wise reprojection loss.
We use a moving inlier threshold throughout the training process that starts loose, and gets more restrictive as training progresses.
Therefore, the network can focus on predictions that are already good, and neglect less precise predictions that would be filtered by RANSAC during pose estimation.
Our training loss is based on the pixel-wise reprojection loss of DSAC* \cite{Brachmann2021dsacstar}:
\begin{equation}
\label{eq:robustloss}
    \ell_{\proj}[\pos_i, \crd_i, \pose_i^*]= 
\begin{cases}
    \hat{e}_{\proj}(\pos_i, \crd_i, \pose^*_i)   & \text{if } \crd_i \in \mathcal{V}\\
    ||\crd_i - \bar{\crd}_i||_0               & \text{otherwise}.
\end{cases}
\end{equation}
This loss optimizes a robust reprojection error $\hat{e}_{\proj}$ for all \emph{valid} coordinate predictions $\mathcal{V}$.
Valid predictions are between 10cm and 1000m in front of the image plane, and have a reprojection error below 1000px. 
For invalid predictions, the loss optimizes the distance to a dummy scene coordinate $\bar{\crd}_i$ that is calculated from the ground truth camera pose assuming a fixed image depth of 10m.
The main difference between DSAC* and our approach is in the definition of the robust reprojection error $\hat{e}_{\proj}$.
DSAC* uses the reprojection error $e_{\proj}$ up to a threshold $\tau$, and the square root of the reprojection error beyond.
Instead, we use $\tanh$ clamping of the reprojection error:
\begin{equation}
    \hat{e}_{\proj}(\pos_i, \crd_i, \pose^*_i) = \tau(t)~\tanh\left(\frac{e_{\proj}(\pos_i, \crd_i, \pose^*_i)}{\tau(t)}\right)
\end{equation}
We dynamically re-scale the $\tanh$ according to a threshold $\tau$ that varies throughout training:
\begin{equation}
    \tau(t) = w(t)~\tau_\text{max} + \tau_\text{min},~\text{with}~~w(t)=\sqrt{1-t^2},
\end{equation}
where $t \in (0,1)$ denotes the relative training progress.
This curriculum implements a circular schedule of threshold $\tau$, which remains close to $\tau_\text{max}$ in the beginning of training, and declines towards $\tau_{\text{min}}$ at the end of training.

\subsection{Backbone Training}
\label{sec:encoder}

As backbone, we can use any dense feature description network \cite{revaud2019r2d2, Tyszkiewicz2020DISK, liu2011siftflow, detone18superpoint}.
However, existing solutions are often optimized towards sparse feature matching. 
Their descriptors are meant to be informative at key points.
In contrast, we need descriptors that are distinctive for any position in the input image.
Thus, we present a simple way to train a feature description network tailored towards scene coordinate regression.
We adhere to the network architecture of DSAC* \cite{Brachmann2021dsacstar}. 
We use the early convolutional layers as our backbone, and split off the subsequent MLP as our scene-specific regression head.
To train the backbone, we resort to the image-level training of DSAC* \cite{Brachmann2021dsacstar} (\cf Fig.~\ref{fig:system_dsac}) but couple it with our training curriculum of Eq.~\ref{eq:robustloss}.

Instead of training the backbone with one regression head for a single scene, we train it with $N$ regression heads for $N$ scenes, in parallel.
This bottleneck architecture forces the backbone to predict features that are useful for a wide range of scenes.
We train the backbone on 100 scenes from ScanNet \cite{dai2017scannet} for 1 week, resulting in 11MB of weights that can be used to extract dense descriptors on any new scene.
See Appendix \ref{supp:implementation-details} for more details on the training process.

\begin{table*}[h!t]
\centering
\footnotesize
\begin{tabular}{clcccccccc}
\toprule
\multicolumn{1}{l}{} & & & & & \multicolumn{2}{c}{7 Scenes} & \multicolumn{1}{l}{} & \multicolumn{2}{c}{12 Scenes} \\
\cmidrule{6-7} \cmidrule{9-10} 
\multicolumn{1}{l}{} & & \multirow{-2}{*}{\begin{tabular}[c]{@{}c@{}}{Mapping w/}\\ {Mesh/Depth}\end{tabular}} & \multirow{-2}{*}{\begin{tabular}[c]{@{}c@{}}{Mapping}\\ {Time}\end{tabular}} & \multirow{-2}{*}{\begin{tabular}[c]{@{}c@{}}{Map}\\ {Size}\end{tabular}} & {SfM poses} & {D-SLAM poses} & & {SfM poses} & {D-SLAM poses} \\
\midrule
 & AS (SIFT) \cite{Sattler2017AS} & \greencell No & \redcell  & \redcell $\sim$200MB & 98.5\% & 68.7\% & & 99.8\% & 99.6\% \\
 & D.VLAD+R2D2 \cite{kapture2020}  & \greencell No & \redcell  & \redcell $\sim$1GB & 95.7\% & 77.6\% & & 99.9\% & 99.7\% \\
 & hLoc (SP+SG) \cite{sarlin2019coarse, sarlin2020superglue} & \greencell No & \redcell  & \redcell $\sim$2GB & 95.7\% & 76.8\% & & 100\% & 99.8\% \\
\multirow{-4}{*}{\rotatebox{90}{{FM}}} 
 & pixLoc \cite{sarlin21pixloc} & \greencell No & \multirow{-4}{*}{\redcell $\sim$1.5h} & \redcell $\sim$1GB & N/A & 75.7\% & & N/A & N/A \\
\midrule
 & DSAC* (Full) \cite{Brachmann2021dsacstar} & \redcell Yes & \redcell 15h & \greencell 28MB & 98.2\% & 84.0\% & & 99.8\% & 99.2\% \\
 & DSAC* (Tiny) \cite{Brachmann2021dsacstar} & \redcell Yes & \redcell 11h & \greencell 4MB & 85.6\% & 70.0\% & & 84.4\% & 83.1\% \\
 & SANet \cite{yang2019sanet} & \redcell Yes & \greencell $\sim$2.3 min & \redcell $\sim$550MB & N/A & 68.2\% & & N/A & N/A \\
\multirow{-4}{*}{\rotatebox{90}{\begin{tabular}[c]{@{}c@{}}{SCR} \\ {(w/ Depth)}\end{tabular}}} 
& SRC \cite{dong2022visual} & \redcell Yes & \greencell 2 min$^\dag$ & \orangecell 40MB & 81.1\% & 55.2\% & & N/A & N/A \\
\midrule\midrule
& DSAC* (Full) \cite{Brachmann2021dsacstar} & \greencell No & \redcell 15h & \orangecell 28MB & {\ul 96.0\%} & \textbf{81.1\%} & & {\ul 99.6\%} & {\ul 98.8\%} \\
& DSAC* (Tiny) \cite{Brachmann2021dsacstar} & \greencell No & \redcell 11h & \greencell 4MB & 84.3\% & 69.1\% & & 81.9\% & 81.6\% \\
\multirow{-3}{*}{\rotatebox{90}{{SCR}}} 
& ACE (ours) & \greencell No & \greencell 5 min & \greencell 4MB & \textbf{97.1\%} & {\ul 80.8\%} & \textbf{} & \textbf{99.9\%} & \textbf{99.6\%} \\
\bottomrule
\end{tabular}
\caption{\textbf{Indoor Relocalization Results.} We report the percentage of frames below a 5cm,5$^\circ$ pose error. Best results in \textbf{bold} for the ``SCR'' group, second best results \underline{underlined}. We list the time needed for mapping, the map size and, whether depth (rendered or measured) is needed for mapping. See the main text and Supp.~for details on these numbers. $^\dag$ does not include time needed to pre-cluster the scene.}
\label{tab:results_indoor}
\vspace{-1em}
\end{table*}

\subsection{Further Improvements}
\label{sec:improvements}

We train the entire network with half-precision floating-point weights. 
This gives us an additional speed boost, especially on budget GPUs.
We also store our networks with float16 precision. 
This allows us to increase the depth of our regression heads while maintaining 4MB maps.
On top of our loss curriculum (see Sec.~\ref{sec:curriculum}), we use a one cycle learning rate schedule \cite{Smith2019cycliclr}, \ie we increase the learning rate in the middle of training, and reduce it towards the end.
We found a small but consistent advantage in overparameterizing the scene coordinate representation:
we predict homogeneous coordinates $\crd'=(x, y, z, w)^\top$ and apply a w-clip, enforcing $w$ to be positive by applying a softplus operation.

\section{Experiments}
\label{sec:experiments}

We implement our approach in PyTorch \cite{paszke2017automatic}, based on the public code of DSAC* \cite{Brachmann2021dsacstar}.
We list our main parameter choices here, and refer to the Appendix \ref{supp:implementation-details} for more details.
We create a training buffer of 8M backbone features in 1 minute, randomly sampled from mapping images.
In the 4 minutes training, we do 16 complete passes over the training buffer with a batch size of 5120.
Our training curriculum starts with a soft threshold of $\tau_{\text{max}}=50\text{px}$ and ends with $\tau_{\text{min}}=1\text{px}$.
We optimize using AdamW \cite{Loshchilov2019DecoupledWD} with a learning rate between $5\cdot10^{-4}$ and $5\cdot10^{-3}$ and a 1 cycle schedule \cite{Smith2019cycliclr}.
We reuse the DSAC* robust pose estimator with 64 RANSAC hypotheses and 10px inlier threshold.
\subsection{Indoor Relocalization}

We conducted experiments on 7Scenes \cite{shotton2013scene} and 12Scenes \cite{valentin2016learning}, two indoor relocalization datasets.
They provide mapping and query images for several small-scale indoor rooms.
In these environments, speedy mapping is particularly desirable.
If mapping takes hours or days, the room might have changed, or the user might have wandered off.
Two sets of ground truth poses are available \cite{brachmann2021limits}, one from running depth-based SLAM \cite{izadi2011kinectfusion, dai2017bundlefusion}, one from running SfM \cite{schoenberger2016sfm}.
We report results on both because relocalizer performance can be biased towards one or the other \cite{brachmann2021limits}.

Our main comparison is to DSAC* \cite{Brachmann2021dsacstar}, our baseline. 
We include other top-performing methods as well, such as feature matching (FM) approaches, and scene coordinate regression (SCR) pipelines that use depth during mapping. 
We list results in Table \ref{tab:results_indoor}.
Multiple approaches achieve high accuracy, but require vastly different resources.
ACE is the only approach that 1) achieves top accuracy 2) in less than 10 minutes per scene 3) with less than 10MB per scene. 
At the same time, it does not need depth -- either measured or rendered from a mesh -- for mapping.

\textbf{Note:} Comparing mapping times and map sizes is a difficult endevour. 
The exact numbers will depend on hardware, implementation details and hyper-parameters.
For our main baseline, DSAC*, we made sure that the numbers are accurate by re-running their public code on our hardware.
For the other approaches, which we mainly list for context, we resort to numbers given in their respective papers, or estimated numbers based on publicly available statistics. 
We detail our reasoning for each method in Appendix \ref{supp:competitors}.
As such, these numbers should be taken with a grain of salt.
But we are confident that the order of magnitude is correct, as confirmed by cross-referencing with other publications.

\begin{table*}[]
\centering
\footnotesize
\begin{tabular}{clccccccccc}
\toprule
\multicolumn{1}{l}{} & & & & & \multicolumn{5}{c}{Cambridge Landmarks} \\
\cmidrule(l){6-10} 
\multicolumn{1}{l}{} & & \multirow{-2}{*}{\begin{tabular}[c]{@{}c@{}}Mapping w/\\ Mesh/Depth\end{tabular}} & \multirow{-2}{*}{\begin{tabular}[c]{@{}c@{}}Mapping\\ Time\end{tabular}} & \multirow{-2}{*}{\begin{tabular}[c]{@{}c@{}}Map\\ Size\end{tabular}} & Court & King's & Hospital & Shop & St. Mary's 
& \multirow{-2}{*}{\begin{tabular}[c]{@{}c@{}}Average \\ (cm / $^\circ$)\end{tabular}} \\ 
\midrule
 & AS (SIFT) \cite{Sattler2017AS} & \greencell No & \redcell  & \redcell $\sim$200MB & 24/0.1 & 13/0.2 & 20/0.4 & 4/0.2 & 8/0.3 & 14/0.2 \\
 & hLoc (SP+SG) \cite{sarlin2019coarse, sarlin2020superglue} & \greencell No & \redcell  & \redcell $\sim$800MB & 16/0.1 & 12/0.2 & 15/0.3 & 4/0.2 & 7/0.2 & 11/0.2 \\
 & pixLoc \cite{sarlin21pixloc} & \greencell No & \redcell  & \redcell $\sim$600MB & 30/0.1 & 14/0.2 & 16/0.3 & 5/0.2 & 10/0.3 & 15/0.2\\
 & GoMatch \cite{zhou2022gomatch} & \greencell No & \redcell  & \orangecell $\sim$12MB & N/A & 25/0.6 & 283/8.1 & 48/4.8 & 335/9.9 & N/A \\
\multirow{-5}{*}{\rotatebox{90}{FM}} 
 & HybridSC \cite{compression2019cvpr} & \greencell No & \multirow{-5}{*}{\redcell $\sim$35min} & \greencell $\sim$1MB & N/A & 81/0.6 & 75/1.0 & 19/0.5 & 50/0.5 & N/A \\
\midrule
 & PoseNet17 \cite{Kendall2017GeometricLF} & \greencell No & \redcell 4 -- 24h & \orangecell 50MB & 683/3.5 & 88/1.0 & 320/3.3 & 88/3.8 & 157/3.3 & 267/3.0\\
\multicolumn{1}{l}{\multirow{-2}{*}{\rotatebox{90}{APR}}}
 & MS-Transformer \cite{Shavit2021MStransformer} & \greencell No & \redcell $\sim$7h & \orangecell $\sim$18MB & N/A & 83/1.5 & 181/2.4 & 86/3.1 & 162/4.0 & N/A \\
\midrule
 & DSAC* (Full) \cite{Brachmann2021dsacstar} & \redcell Yes & \redcell 15h & \orangecell 28MB & 49/0.3 & 15/0.3 & 21/0.4 & 5/0.3 & 13/0.4 & 21/0.3 \\
 & SANet \cite{yang2019sanet} & \redcell Yes & \greencell $\sim$1min & \redcell $\sim$260MB & 328/2.0 & 32/0.5 & 32/0.5 & 10/0.5 & 16/0.6 & 84/0.8\\
\multirow{-3}{*}{\rotatebox{90}{\begin{tabular}[c]{@{}c@{}}SCR \\ w/ Depth\end{tabular}}}
& SRC \cite{dong2022visual} & \redcell Yes & \greencell 2 min$^\dag$ & \orangecell 40MB & 81/0.5 & 39/0.7 & 38/0.5 & 19/1.0 & 31/1.0 & 42/0.7\\
\midrule
\midrule
\multirow{4}{*}{\rotatebox{90}{SCR}}
 & DSAC* (Full) \cite{Brachmann2021dsacstar} & \greencell No & \redcell 15h & \orangecell 28MB & {\ul 34/0.2} & \textbf{18/0.3} & \textbf{21/0.4} & \textbf{5/0.3} & {\ul 15/0.6} & {\ul 19/0.4} \\
 & DSAC* (Tiny) \cite{Brachmann2021dsacstar} & \greencell No & \redcell 11h & \greencell 4MB & 98/0.5 & {\ul 27/0.4} & 33/0.6 & {\ul 11/0.5} & 56/1.8 & 45/0.8\\
& ACE (ours) & \greencell No & \greencell 5 min & \greencell 4MB & {43/0.2} & 28/0.4 & {31/0.6} & \textbf{5/0.3} & {18/0.6} & 25/0.4\\
\cmidrule{2-11}
& Poker (Quad ACE Ensemble) & \greencell  No & \orangecell 20 min & \orangecell 16MB & \textbf{28/0.1} & \textbf{18/0.3} & {\ul 25/0.5} & \textbf{5/0.3} & \textbf{9/0.3} & \textbf{17/0.3} \\
\bottomrule
\end{tabular}
\caption{\textbf{Cambridge Landmarks \cite{kendall2015posenet} Results.} We report median rotation and position errors. Best results in \textbf{bold} for the ``SCR'' group, second best results \underline{underlined}. Methods using depth for mapping rely on a dense multi-view stereo mesh that took hours to compute \cite{kendall2015posenet, Brachmann2018dsacpp}. See the main text and Supp.~for details about mapping times and storage demands. $^\dag$ does not include time needed to pre-cluster the scene.}
\label{tab:results_cam}
\end{table*}

\subsection{Outdoor Relocalization}

\paragraph{Cambridge Landmarks \cite{kendall2015posenet}.} 
This dataset collects mapping and query images of buildings across the old town of Cambridge.
Ground truth poses stem from reconstructing mapping and query images jointly using SfM \cite{wu2011visualsfm}.
We report our results in Table \ref{tab:results_cam}.
Feature matching-approaches do very well on this dataset, presumably due to their similarity to the SfM reference algorithm \cite{brachmann2021limits}.
ACE performs reasonably well compared to DSAC* considering the spatial extent of each scene and our 4MB memory footprint.
We provide a variant of ACE, denoted \emph{Poker}, where we split a scene into four ACE models, and return the pose estimate with the largest inlier count at query time (see Supp.~for details).
\emph{Poker} outperforms DSAC* on average while still being comparatively lean \wrt mapping time and storage.

\begin{figure*}[t]
  \centering
   \includegraphics[width=1.0\linewidth]{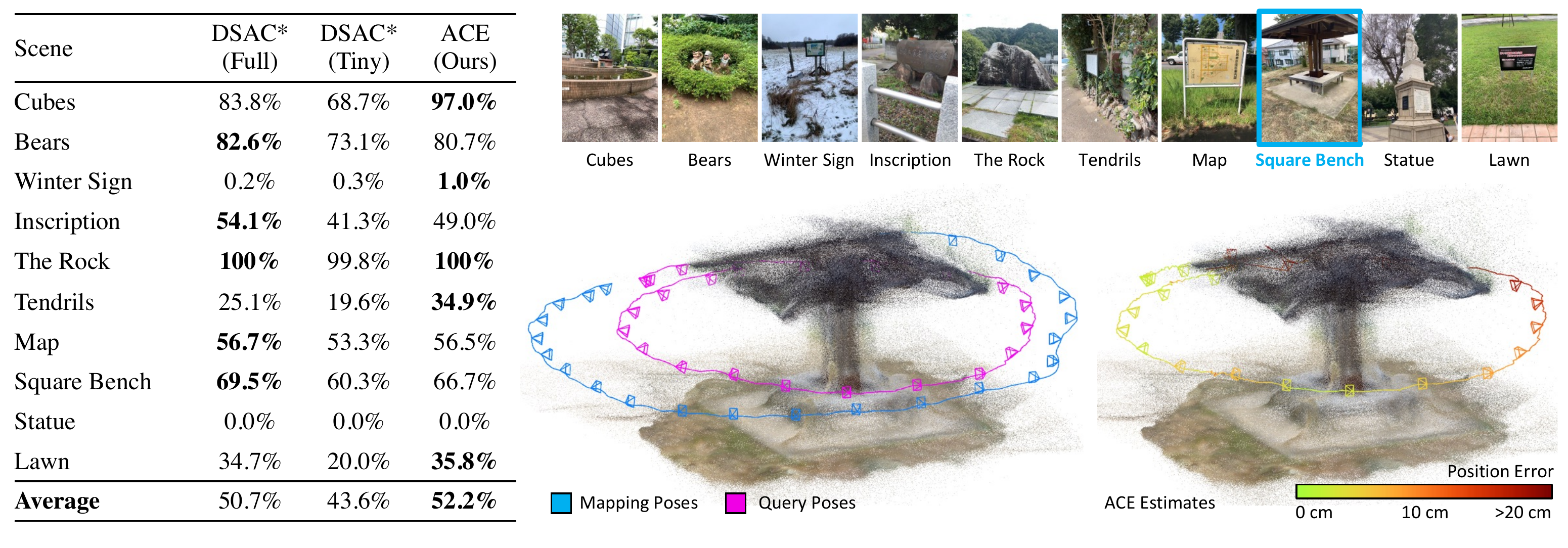}
   \caption{\textbf{The Wayspots Dataset.} We curate a new relocalization dataset from a public corpus of phone scans \cite{arnold2022mapfree}. (left) We show accuracy as the percentage of frames below 10cm,5$^\circ$ pose error. Best results in \textbf{bold}. (right) One mapping image of each scene, and a visualisation of ground truth poses and ACE estimates for one scene. The dataset does not provide 3D point clouds, so we show the ACE map instead.
   See Appendix \ref{supp:map_vis} for details of the process we use to render 3D point clouds from the trained ACE maps.
   }
   \label{fig:mapfree}
\end{figure*}

\begin{figure*}[t]
  \centering
   \includegraphics[width=1.0\linewidth]{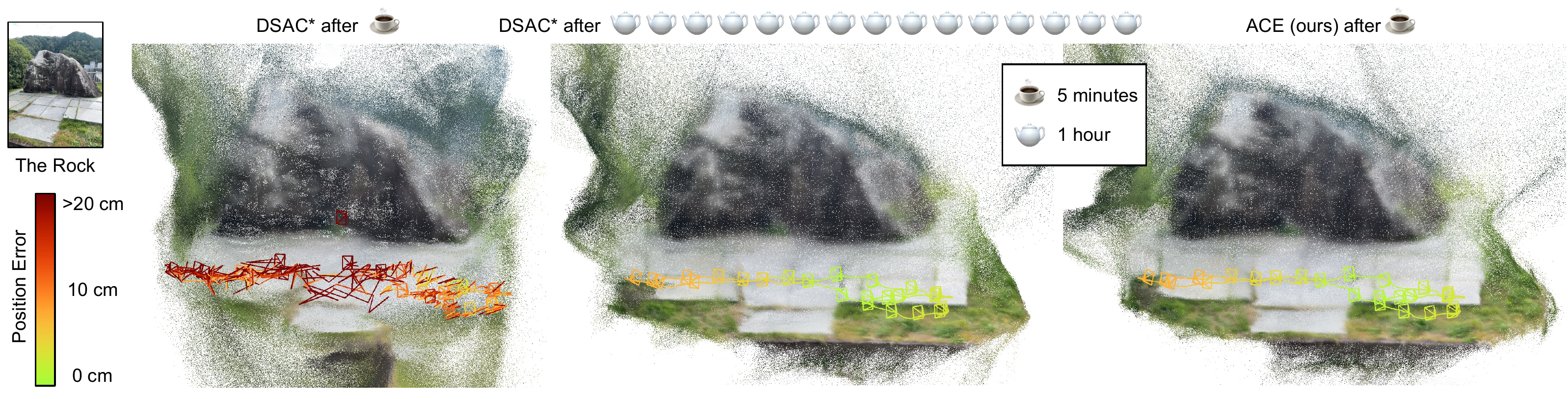}
   \caption{\textbf{Qualitative results.} We compare qualitative results of ACE and DSAC* on one scene of the Wayspots dataset. Full training of DSAC* takes 15 hours. When stopped after 5 minutes, relocalization accuracy is poor. We show the learned map for each method.}
   \label{fig:second_teaser}
\end{figure*}

\paragraph{Wayspots.}
The previous datasets are not ideal for fully showcasing the advantages of our method.
Their mapping poses either stem from D-SLAM, where depth would be available, or SfM reconstruction, where 3D point clouds would be available.
However, RGB-based visual odometry runs on millions of phones \cite{arcore, arkit} providing mapping images and poses, but neither depth nor full 3D point clouds with feature descriptors.
Therefore, we curate a new relocalization dataset, denoted \emph{Wayspots}, from a publicly available corpus of phone scans.
In particular, we select 10 consecutive scenes from the training split of the MapFree dataset \cite{arnold2022mapfree}.
Each scene depicts a small outdoor place and comes with two full, independent scans. 
We use one for mapping and one for query. 
Arnold \etal \cite{arnold2022mapfree} reconstructed ground truth poses of both scans using SfM \cite{schoenberger2016sfm}.
We register the original phone trajectories to the SfM poses.
In our experiments, we used the original phone trajectories for mapping, and SfM poses solely for evaluation.
We refer to Appendix \ref{supp:wayspots} for details on the dataset and its curation.
We show an overview of all scenes, and our main results in Fig.~\ref{fig:mapfree}.
We outperform DSAC* on average while being two orders of magnitude faster in mapping.
In Fig.~\ref{fig:second_teaser}, we show a qualitative comparison of ACE and DSAC* on the ``Rock'' scene, including a variant of DSAC* that was stopped after 5 minutes mapping.

\subsection{Analysis}
\begin{figure}[h!]
  \centering
   \includegraphics[width=1.0\linewidth]{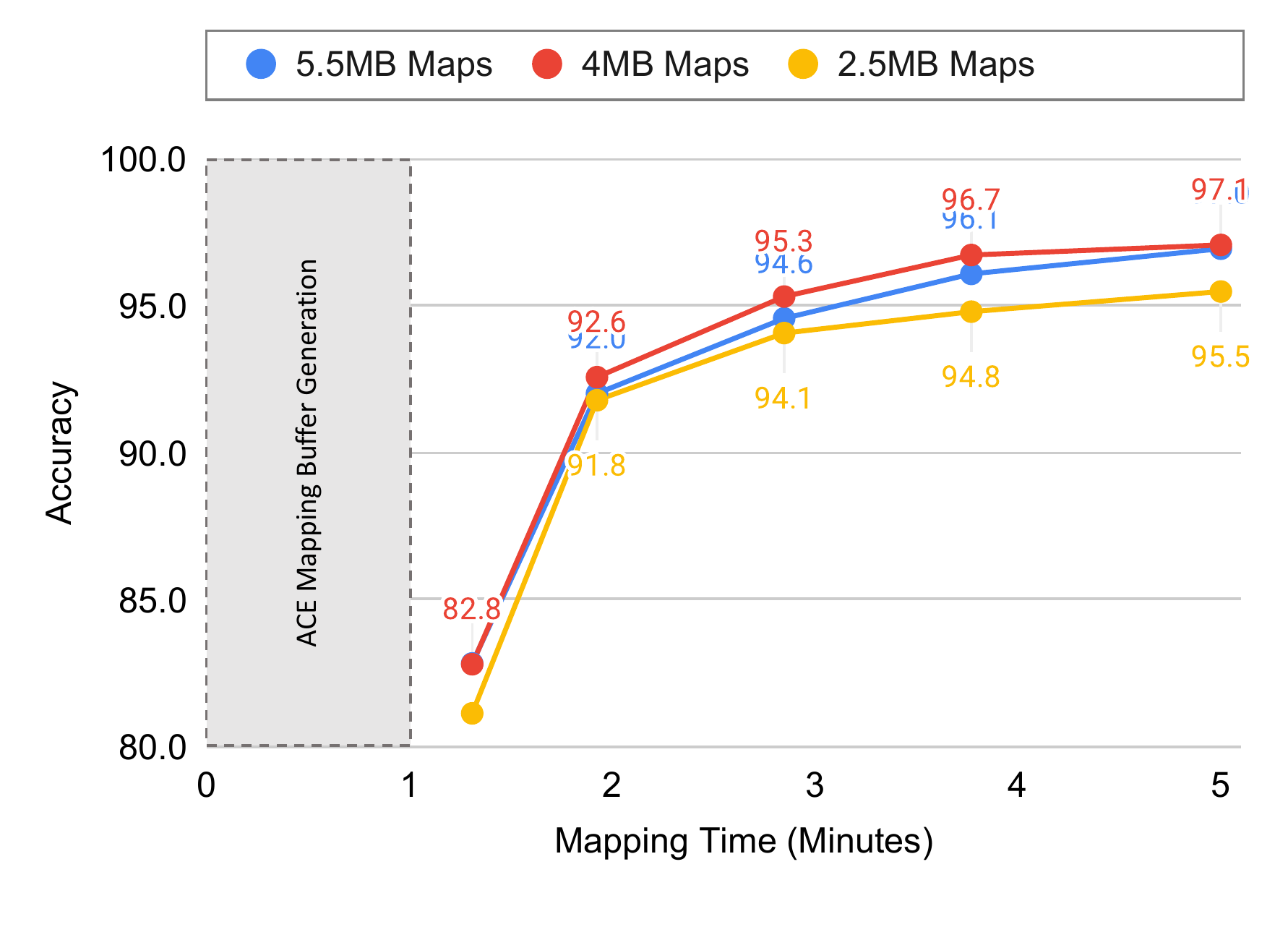}
   \vspace{-0.65cm}
   \caption{\textbf{Map Size.} We vary the map size of ACE by using more or less MLP layers. 4MB maps correspond to our standard settings. Results on 7Scenes with SfM pseudo ground truth poses. }
   \vspace{-0.35cm}
   \label{fig:time}
\end{figure}

\paragraph{Mapping Time.}
The speed of training a network depends on the GPU model.  
In Table \ref{tab:mapping_time}, we compare mapping times of ACE and DSAC* on a premium GPU (NVIDIA V100) with a more affordable model (NVIDIA T4)\footnote{At time of writing, instantiating a virtual machine with a Tesla T4 on a popular cloud computing provider~\cite{gcpgpupricing} costs USD 0.35/h, whereas the more powerful Tesla V100 GPU costs USD 2.48/h, 7x more expensive.}.
For DSAC*, mapping takes twice as long on the cheaper GPU.
The difference is smaller for the ``Tiny'' version of DSAC* that uses a leaner network, but mapping still takes hours.
Conversely, ACE experiences only a 10\% slowdown on the T4 GPU, due to having to train just the regression head, in half-precision. 

\begin{table}[]
\centering
\footnotesize
\begin{tabular}{llcc}
\toprule
GPU                   & Method       & Mapping Time & ACE Speed-up \\ \midrule

\multirow{3}{*}{NVIDIA V100} & ACE          & 291s         & -            \\
                      & DSAC* (Tiny) & 11h          & 130x          \\
                      & DSAC* (Full) & 15h          & 180x         \\ \midrule

\multirow{3}{*}{NVIDIA T4}   & ACE          & 327s         & -            \\
                      & DSAC* (Tiny) & 14h          & 150x          \\
                      & DSAC* (Full) & 28h          & 310x         \\ \bottomrule
\end{tabular}
\caption{\textbf{Mapping Times.} Comparison of methods on different GPUs. T4 GPUs offer less compute capabilities but are considerabley cheaper. DSAC* (Full) uses a 28MB network. DSAC* (Tiny) uses a smaller, 4MB network.}
\label{tab:mapping_time}
\end{table}

\paragraph{Map Size.}
The network architecture of our regression head determines the size of our maps.
In particular we can vary the number of layers in our MLP, see Fig.~\ref{fig:time} for an analysis.
Smaller maps of 2.5MB achieve slightly lower accuracy, but still yield comfortable relocalization rates above 95\% on 7Scenes.
Using larger maps of 5.5MB does not pay off: because of their increased size, these networks undergo fewer passes over the training buffer in 5 minutes.
Note how ACE achieves $\sim$80\% accuracy after a single epoch (75s).

\paragraph{More Experiments in the Appendix.} 
We demonstrate the positive impact of gradient de-correlation (Appx.~\ref{supp:exp:decorrelation}), predicting homogeneous scene coordinates and our loss curriculum (Appx.~\ref{supp:loss}).
Furthermore, we substitute the ACE feature backbone with various off-the-shelf networks, including SuperPoint \cite{detone18superpoint} and DISK \cite{Tyszkiewicz2020DISK}, to find them  less suited for our task (Appx.~\ref{supp:exp:backbone}).
We also vary the dimensionality of ACE backbone features to find more dimensions working better.

\section{Conclusion and Future Work}

We have presented ACE, a relocalizer able to map new environments in 5 minutes. 
ACE reduces cost and energy consumption of mapping by two orders of magnitude compared to previous RGB-based scene coordinate regression approaches, making this family of algorithms practical. 

The changes from previous state-of-the-art relocalizers that we propose in this paper are mainly conceptual, leveraging decorrelation of gradients by patch-level training.
We see further potential for speedups by clever engineering, such as interleaving buffer creation and training in separate threads, or early stopping for easy scenes.

\paragraph{Acknowledgements:} We thank Torsten Sattler for helpful comments about the resource footprint of feature-based relocalizers, Mohamed Sayed for help with the 3D visualisation of ACE, and Cl\'{e}ment Godard for early contributions to the code base of ACE.

\appendix

\section{Implementation Details}
\label{supp:implementation-details}

\subsection{Backbone Training}

Our backbone architecture consists of the first 10 layers (including skip connections) of the DSAC* \cite{Brachmann2021dsacstar} network design.
The backbone takes a gray scale image as input, and successively decreases the spatial resolution to $\frac{1}{8}$ while increasing the channel dimensions to 512.

We train our backbone on 100 scenes in parallel, attaching 100 regression heads to it.
Each regression head is a multi-layer perceptron with 6 layers and width 512.
There is a skip connection after the first 3 layers of each head.
We train the backbone with half-precision floating point weights.
We apply strong data augmentation to the mapping images during backbone training.
We use a brightness and contrast jitter of 40\%; a saturation and hue jitter of 30\% (before converting to grayscale); and we randomly re-scale the images between 240px and 960px height.
We use an inverse scale sampling strategy, \ie we rescale images according to $\frac{1}{s}$ where $s$ is a uniformly random scale factor.
This mimics scale changes caused by the camera uniformly moving towards a scene.
We warp images using homographies that correspond to random 3D rotations of up to 40$^\circ$.

We train the backbone with a batch size of 6 images per regression head. 
To avoid memory issues, we compute forward and backward passes for 10 regression heads at a time.
We accumulate gradients for all 100 regression heads before triggering a parameter update.
We train on the first 100 training scenes of ScanNet \cite{dai2017scannet}, \ie scenes \emph{scene0000} to \emph{scene0099}.
ScanNet provides multiple scans per scene, but they are not aligned. 
Thus, we use only one (the first) scan for each scene, \ie scans \emph{scene0000\_00} to \emph{scene0099\_00}.

\subsection{ACE Head Training}

\subsubsection{Network Design}
\label{sec:network-design}

The multi-layer perceptron we use as scene coordinate regression head is composed of 8 1x1 convolutional layers, all of width 512, with skip connections after layer 3 and 6; followed by a final 1x1 convolutional layer producing the scene coordinates.
All the head layers use half-precision floating point weights.
As mentioned in the main text, we experimented with both directly regressing 3D scene coordinates (in which case the last layer would output a 3-channel tensor), and regressing homogeneous coordinates.
In the latter case, the last layer outputs a 4D tensor of $(\dot{x}, \dot{y}, \dot{z}, \hat{w})$, with $\mathbf{\dot{y}} = (\dot{x}, \dot{y}, \dot{z})^T$ being the homogeneous representation of the 3D scene coordinates, and $\hat{w} \in \mathbb{R}$ being an un-normalized homogeneous parameter.
Before de-homogenizing the scene coordinates we compute $w \in \mathbb{R}^+$ from $\hat{w}$ applying a biased (and clipped) Softplus operator to ensure that the final 3D scene coordinates would be valid and usable in the following RANSAC step.

Specifically, we compute $w$ as follows:
\begin{equation}
    w = \min{\left( \frac{1}{\mathit{S}_{\mathit{min}}}, \beta^{-1} \cdot \log{\left( 1 + \exp{\left( \beta \cdot \hat{w} \right)} \right)}  + \frac{1}{\mathit{S}_{\mathit{max}}}\right)}
\end{equation}
Where $\mathit{S}_{\mathit{min}}$ and $\mathit{S}_{\mathit{max}}$ are the values we use to clip the scale factor determined by $w$ (in our experiments we set them to $0.01$ and $4.0$ respectively), and $\beta$ is the parameter used to ensure that when the network outputs $\hat{w} = 0$, the resulting homogeneous parameter $w$ becomes $1$.
This is to steer the network towards producing a ``neutral'' homogeneous parameter (centered on $1$).
As such, for our experiments, we set $\beta = \frac{\log(2)}{1 - \mathit{S}_{\mathit{max}}^{-1}}$.

We then de-homogenize the output of the network into the tensor $\mathbf{y}$ containing 3D scene coordinates, as usual:
\begin{equation}
    \mathbf{y} = \frac{\mathbf{\dot{y}}}{w}
\end{equation}

In both cases the coordinates output by the network, for numerical stability, are learned relatively to the ``mean'' translation of the camera poses associated to the mapping frames.
The final step of the scene coordinate regression process is then to add the mean back to the prediction, before passing the tensor of 3D scene coordinates to the PnP-RANSAC stage.

\subsubsection{Training Details}
As mentioned in the main text, key to the speed with which we are able to train an ACE map is the fact that, first, we pre-fill a buffer of patch features extracted from the set of training images (using the scene-agnostic backbone, this takes $\sim$1min), and then we just need to optimize the weights of the regression head.

More specifically, we allocate a buffer able to contain 8 million 512-channel patch descriptors, together with their associated 2D location in the source image, mapping camera pose, and intrinsic parameters.
We fill the buffer by repeatedly cycling over the shuffled training sequence.
Each image is augmented using a similar approach as the backbone training, but with a different set of hyperparameters that we detail next.
We apply a brightness and contrast jitter of 10\%, but no color and saturation jitter; randomly rescale the input images between 320px and 720px height; and, finally, we apply in-plane random rotations of up to 15$^\circ$.
We can use these weaker data augmentation parameters to still train the ACE regression heads accurately because the backbone has already been trained on strongly augmented images.
The training images are passed through the convolutional backbone and, for each one, we randomly select 1024 patches and their corresponding feature descriptors, which we then copy into the training buffer, together with the other metadata (2D patch location, camera pose and intrinsics).

We train the ACE head by repeatedly iterating over the shuffled training buffer, as described in the main text.
Specifically, for a 5 minute mapping time (including the time spent filling the buffer), we do this 16 times (epochs).

\subsubsection{Choice of Loss Function}
\label{supp:loss}

\begin{table*}[t!]
    \centering
    \begin{tabular}{lccccc}
    \toprule
    \multirow{3}{*}{Scene} & \multirow{3}{*}{\begin{tabular}{c}$\mathit{L1} \rightarrow \sqrt{\mathit{L1}}$ \\ (DSAC*\cite{Brachmann2021dsacstar}) \end{tabular}} & \multicolumn{4}{c}{$\tanh$-based} \\
    \cmidrule{3-6}
    & & \multirow{2}{*}{\begin{tabular}{c}fixed \\ $\tau = 50px$ \end{tabular}} & \multirow{2}{*}{\begin{tabular}{c}linear schedule \\ $\tau = 50 \rightarrow 1px$ \end{tabular}} & \multirow{2}{*}{\begin{tabular}{c}circular schedule \\ $\tau = 50 \rightarrow 1px$ \end{tabular}} & \multirow{2}{*}{\begin{tabular}{c}circular schedule \\ + homogeneous coords \end{tabular}} \\
    \\
    \midrule
    Chess & 100 & 100 & 100 & 100 & 100 \\
    Fire & 99.1 & 99.3 & 99.5 & 99.3 & 99.5 \\
    Heads & 99.8 & 99.8 & 99.8 & 99.7 & 99.7 \\
    Office & 99.5 & 99.6 & 99.5 & 99.7 & 100 \\
    Pumpkin & 100 & 99.9 & 99.8 & 99.8 & 99.9 \\
    RedKitchen & 96.7 & 97.0 & 97.0 & 98.3 & 98.6 \\
    Stairs & 69.4 & 68.1 & 66.8 & 70.5 & 81.9 \\
    \midrule
    Average & 94.9 & 94.8 & 94.6 & 95.3 & 97.1 \\
    \bottomrule
    \end{tabular}
    \caption{\textbf{Loss Functions.} Performance of ACE on the 7 Scenes dataset \cite{shotton2013scene} when trained using different loss functions. For each scene we report the \%-age of frames localized within 5cm/5$^\circ$ pose error. All columns except the right-most one deploy an ACE regression head directly predicting 3D scene coordinates (see Sec.~\ref{sec:network-design}). The last column shows the contribution of the homogeneous coordinate encoding, which is helpful, especially in a complex scene like ``Stairs''. As such, the results in the main text use the $\tanh$-based loss with circular schedule and homogeneous coordinates.}
    \label{tab:loss-ablation}
\end{table*}

In order to train the ACE head networks to regress accurate 3D scene coordinates, we use a $\tanh$-based loss on the reprojection errors, dynamically rescaled according to a circular schedule with a threshold decreasing throughout the length of the training process (see Sec.~3.2 of the main text for details).
We experimented with alternative loss functions as well, before settling on the one used in the main text, and in Table~\ref{tab:loss-ablation} we show their accuracy on the 7 Scenes dataset~\cite{shotton2013scene}.

Firstly, we experimented with the reprojection loss described in the DSAC* paper~\cite{Brachmann2021dsacstar}.
This loss is computed piecewisely, using the L1 norm of the reprojection error up to a threshold of $\tau=100$px, and switching to the square root of the L1 norm past it.
As described in the main text, we also experimented with $\tanh$ losses, which effectively clamp each patch's individual contribution to the overall error used for optimization during training.
The right side of Table~\ref{tab:loss-ablation} shows the accuracy for different variants of the $\tanh$ loss: (a) using a constant $\tau$ threshold set at 50px throughout the training process; (b) \textit{linearly} decreasing $\tau$ from 50px to 1px, \ie the $w(t)$ in Eq.~8 of the main text computed as: $w(t) = 1 - t$; (c) \textit{quadratically} decreasing $\tau$ from 50px to 1px (the circular schedule, as in Eq.~8 of the main text); and, finally, (d) applying the circular schedule as in (c) but also using the homogeneous coordinate prediction, as described in the previous section (note that all other columns of the table were directly predicting 3D coordinates).

Overall we see that using the circular schedule, with a threshold decreasing over time, in order to focus the network training to improve the prediction of reliable scene coordinates (with the knowledge that unreliable predictions will be effectively filtered by the RANSAC stage during localization) allows us to improve from the results obtained using DSAC*'s piecewise loss.
We also see how the overparametrized homogeneous representation consistently improves the overall accuracy of the method, especially on difficult scenes such as ``RedKitchen'' and ``Stairs''.
The latter is the variant we chose to deploy throughout the paper.

\subsection{Ensemble Variant}
\label{supp:ensemble}

In the main text, we introduced an \textit{ensemble} variant of ACE that we evaluated on the five scenes of the Cambridge Landmarks dataset~\cite{kendall2015posenet}.
We named that variant \textit{``Poker''}, as it is composed of four identical but independently trained network heads, collaboratively contributing to the estimation of the camera poses in the larger environments part of the dataset.
The main reasoning behind the introduction of the ensemble of networks is that some of the scenes of the dataset are covering a large area around the landmarks in the city of Cambridge.

We postulated that, instead of training a single ACE head to regress 3D scene coordinates for the entirety of each landmark (a hard task, considering the map size and training times we target), we could instead cluster the mapping frames into sub-regions and deploy multiple regression heads, each focusing on a specific area of the scene.

In the following paragraphs, we describe how we split the training data and ran localization to produce the results we showed in Table~2 of the main text.
We also present additional results detailing how the ensemble variants perform with a varying number of independently trained regression heads.

\subsubsection{Ensemble Training}

Previous works~\cite{Brachmann2019ESAC} showed how using a \textit{Mixture of Experts} can help improving the performance of large-scale outdoor camera localization.
In this paper we adopt an approach inspired by the ESAC paper~\cite{Brachmann2019ESAC}, training multiple ACE heads on independent subsets of frames part of the training split of each scene.
We adapt the code they publicly released in order to spatially cluster the images according to the position of the camera at capture time.
This is slightly different from the approach described in the ESAC paper, as they used the median scene coordinate for each image -- which is available as part of the Cambridge Landmarks training set.
We decided not to use that information in order to make the method more resembling of a realistic use-case, where the only information available at mapping time would be the pose of the camera (provided by ARKit\cite{arkit}/ARCore\cite{arcore}), but not the 3D point cloud of the scene (which would instead have to be reconstructed offline via SfM).

Specifically, given a set of posed RGB images $\mathcal{I}_\text{M}$ to use as training to map a large-scale environment, we apply hierarchical clustering to the translation component of each frame's pose in order to obtain $N$ disjoint training sets $[ \mathcal{I}_{\text{M}_1}..\mathcal{I}_{\text{M}_N} ]$. 
Clustering is performed as follows:
\begin{itemize}[noitemsep]
    \item Initialization: Put all images in the first cluster $\mathcal{I}_{\text{M}_1}$.
    \item While the target number of clusters $N$ has not been reached:
    \begin{itemize}[noitemsep]
        \item Select the largest cluster (by number of images associated to it).
        \item Split it into two clusters using kMeans~\cite{kmeans++} (The input to kMeans is the translation component of the camera pose).
        \item Replace the large cluster with the two new ones output by kMeans.
    \end{itemize}
    \item Return the $N$ clusters $[ \mathcal{I}_{\text{M}_1}..\mathcal{I}_{\text{M}_N} ]$.
\end{itemize}

After clustering, we train $N$ ACE regression heads $f_{H_i}$, one for each set of training images.
As each head is trained independently from each other, both the mapping time and the map size scale linearly with the number of elements part of the ensemble, although the process can be trivially parallelized to save time by using multiple GPUs, if available.

\subsubsection{Ensemble Localization}

Localization for the ensemble variant of ACE is straightforward: we simply run the standard localization pipeline separately for each trained map, then pick the camera pose having the most inliers after RANSAC and LM-based refinement.

It's worth noting that, since the feature extraction backbone is scene-agnostic, we can optimize the localization phase by running the query images through the backbone $f_B$ just once, outputting image features $\mathbf{f}_i$.
We can then pass such features through the $N$ ACE coordinate regression heads $f_{H_i}$ to regress the scene coordinates, which are then passed to $N$ instances of the RANSAC algorithm for pose estimation.

\subsubsection{Ensemble Results}

In Table~\ref{tab:ensemble-results} we show how the performance of ACE on the Cambridge Landmarks dataset~\cite{kendall2015posenet} varies when changing the number of maps part of the ensemble.
Note how the error initially decreases while we increase the number of ACE heads being trained -- hinting to the fact that limiting the spatial extent covered by the frames used for training is advantageous with such a fast/aggressive training regime -- then saturates past the ensemble with 4 ACE regressors.
The fact that the ``Shop'' scene does not seem to benefit from using multiple ACE heads for regression can be explained by considering that the mapped area is already limited, thus not requiring further clustering.
We chose to present the results for an ensemble of four maps (which we named \textit{``Poker''}) in the main text because it achieves better results than the baseline method (DSAC*) within a fraction of the mapping time.

The clustering and localization approach described in this section has been deliberately kept simple to showcase the potential of using multiple regression heads to map large areas (as this is not the focus of this work, but merely an extension) but can, with clever engineering and tuning, be made faster and more accurate, for example by tailoring the size of each ensemble to the area covered by its scene.

\begin{table*}[t!]
    \centering
    \begin{tabular}{lcccccccc}
    \toprule
         \multirow{2}{*}{Scene} & \multirow{2}{*}{\begin{tabular}[c]{@{}c@{}}{DSAC*}\\ {(Full)\cite{Brachmann2021dsacstar}}\end{tabular}} & \multirow{2}{*}{ACE} & \multicolumn{6}{c}{ACE Ensemble} \\
         \cmidrule{4-9}
         & & & 2 models & 3 models & 4 models & 5 models & 6 models & 7 models \\
         \midrule
         Court & 34/0.2 & 43/0.2 & 32/0.2 & 32/0.2 & 28/0.1 & 29/0.1 & 28/0.1 & 27/0.1 \\
         King's & 18/0.3 & 28/0.4 & 23/0.4 & 20/0.4 & 18/0.3 & 18/0.3 & 18/0.3 & 16/0.3 \\
         Hospital & 21/0.4 & 31/0.6 & 22/0.4 & 23/0.5 & 25/0.5 & 24/0.5 & 25/0.5 & 23/0.5 \\
         Shop & 5/0.3 & 5/0.3 & 6/0.2 & 5/0.3 & 5/0.3 & 5/0.3 & 6/0.3 & 5/0.3 \\
         St. Mary's & 15/0.6 & 18/0.6 & 13/0.4 & 12/0.4 & 9/0.3 & 9/0.3 & 9/0.3 & 9/0.3 \\
         \midrule
         Average (cm/$^\circ$) & 19/0.4 & 25/0.4 & 19/0.3 & 18/0.4 & 17/0.3 & 17/0.3 & 17/0.3 & 16/0.3\\
         \midrule
         Mapping Time & \redcell 15h & \greencell 5min & \greencell 10min & \orangecell 15min & \orangecell 20min & \orangecell 25min & \redcell 30min & \redcell 35min \\
         Map Size (MB) & \orangecell 28MB & \greencell 4MB & \greencell 8MB & \orangecell 12MB & \orangecell 16MB & \orangecell 20MB & \orangecell 24MB & \orangecell 28MB \\
         \bottomrule         
    \end{tabular}
    \caption{\textbf{ACE Ensemble.} Performance of the ACE Ensemble variants on scenes from the Cambridge Landmarks dataset~\cite{kendall2015posenet}. We report the median position and rotation errors, together with the time and storage required for each map.}
    \label{tab:ensemble-results}
\end{table*}

\section{Experimental Details}
\label{sec:experimental-details}

\subsection{Resource Footprint of Competitors}
\label{supp:competitors}

\subsubsection{Active Search \cite{sattler2012improving, Sattler2017AS}}

\paragraph{Mapping Time.} 
For 7Scenes \cite{shotton2013scene} and 12Scenes \cite{valentin2016learning}, we report the average triangulation time of COLMAP \cite{schoenberger2016sfm} using SIFT \cite{sift} features and known camera poses.
These timings were kindly provided by the authors of \cite{brachmann2021limits} who generated alternative SfM ground truth for the aforementioned datasets.
They ran feature detection and parts of feature matching on GPU, while triangulation ran on CPU. 
Reconstruction times vary a lot between scenes:
\emph{Heads} is the fastest scene with 6 minutes reconstruction, \emph{RedKitchen} is the slowest scene with 10 hours reconstruction.
On average, 7Scenes took 3.3h per scene to reconstruct, and this is the time we report for Active Search in Fig.~1 of the main paper.
12Scenes reconstructed in 35 minutes on average per scene, due to having fewer mapping frames than 7Scenes.
Averaging the reconstruction times across all scenes of 7Scenes and 12Scenes yields 1.5h, the number we give in Table 1 of the main paper.

For Cambridge Landmarks \cite{kendall2015posenet}, we found no public record of reconstruction times. 
The SfM pseudo ground truth was generated by running VisualSfM \cite{wu2011visualsfm} on the combined set of training and test images.
It is unclear whether running SfM on the training set alone, as one would have to do in practise, would result in camera poses of comparable quality.
In terms of mapping frame count, we found Cambridge Landmarks comparable to 12Scenes on average. 
Thus, we assume that the 35 minutes reconstruction time for 12Scenes would approximately also hold for Cambridge Landmarks, and report this number in Table 2 of the main paper.

Note: These times do not include any scene-specific pre-processing that the relocalizer might do on top of the SfM reconstruction, such as building acceleration data structures for faster matching.

\paragraph{Map Size.} 
For each scene, Active Search needs to store the 3D point cloud, a list of feature descriptors per 3D point, as well as co-visibility information and a visual dictionary.
Overall, feature descriptors constitute the largest portion of the map size, and we disregard all other factors. 
Active Search stores SIFT \cite{sift} descriptors, \ie 128 byte if stored in unsigned char precision.
At most, Active Search could store all observed descriptors for each 3D point in the SfM reconstruction, \ie one descriptor for all feature detections that lead to a triangulated 3D point.
However, in practise, Active Search runs a clustering algorithm on all descriptors per 3D point, and creates one representative descriptor per cluster.
Based on communication with the authors of \cite{Sattler2017AS}, we assume that Active Search stores descriptors for only 30\% of feature observations.
The COLMAP reconstruction statistics reported in \cite{brachmann2021limits} include the number of triangulated 3D points, and the number of feature observations for 7Scenes\cite{shotton2013scene} and 12Scenes\cite{valentin2016learning}.
We use these statistics to calculate an average descriptor storage demand of 200MB per scene for these two datasets.
This is the number we give in Table 1 of the main paper.

For Cambridge Landmarks \cite{kendall2015posenet}, Camposeco \etal report an average memory demand of 200MB per scene in \cite{compression2019cvpr} for Active Search.
This is the number we give in Table 2 of the main paper.

\subsubsection{D.VLAD+R2D2 \cite{revaud2019r2d2}}

\paragraph{Mapping Time.}
We report the average triangulation time of COLMAP \cite{schoenberger2016sfm} in Table 1 of the main paper.
Please see our reasoning for Active Search, explained above. 
This assumes that a COLMAP reconstruction using R2D2 \cite{revaud2019r2d2} takes approx.~the same time as when using SIFT \cite{sift}.

\paragraph{Map Size.}
Following our reasoning for Active Search (see above) we approximate the map size with the memory needed to store descriptors.
We assume that D.VLAD+R2D2 stores descriptors for all feature observations, and uses 256 bytes per descriptor for 128 floating-point entries with half precision.
We found no public record of R2D2 reconstruction statistics for 7Scenes\cite{shotton2013scene} and 12Scenes\cite{valentin2016learning}.
Re-using the reconstruction statistics of \cite{brachmann2021limits}, we arrive at 1GB map size on average.
This is the number we give in Table 1 of the main paper.

\subsubsection{hLoc (SP+SG) \cite{sarlin2019coarse, sarlin2020superglue}}

\paragraph{Mapping Time.}
We report the average triangulation time of COLMAP \cite{schoenberger2016sfm} in Table 1 and Table 2 of the main paper.
Please see our reasoning for Active Search, as explained above. 
This assumes that a COLMAP reconstruction using SuperPoint \cite{detone18superpoint} takes approximately the same time as when using SIFT \cite{sift} features.

\paragraph{Map Size.}

Following our reasoning for Active Search (see above) we approximate the map size with the memory needed to store descriptors.
We assume that hLoc stores descriptors for all feature observations, and uses 512 bytes per descriptor for 256 floating-point entries with half precision.
We found no public record of SuperPoint reconstruction statistics for 7Scenes\cite{shotton2013scene} and 12Scenes\cite{valentin2016learning}.
Re-using the reconstruction statistics of \cite{brachmann2021limits}, we arrive at 2GB map size on average across 7Scenes and 12Scenes.
This is the number we give in Table 1 of the main paper.
For 7Scenes alone, we arrive on 3.5GB average map size, and this is the number we state in Fig.~1 of the main paper.
This aligns well with the 3.3GB average map size for hLoc on 7Scenes that is reported by Zhou \etal in \cite{zhou2022gomatch}.
For Cambridge Landmarks \cite{kendall2015posenet}, Zhou \etal report an average memory demand of 800MB per scene in \cite{zhou2022gomatch} for hLoc (SP+SG).
This is the number we give in Table 2 of the main paper.

\subsubsection{pixLoc \cite{sarlin21pixloc}}

\paragraph{Mapping Time.}
pixLoc is a method that works on top of existing 3D map reconstructions.
Thus, as mapping time in Table 1 and Table 2 of the main paper, we report the average triangulation time of COLMAP \cite{schoenberger2016sfm} using SIFT \cite{sift} features, see our reasoning for Active Search above.

\paragraph{Map Size.}
pixLoc relies on dense features to optimize map-to-image alignment.
We assume that it is more advantageous to store the original mapping images and compute features on the fly, than it is to store dense descriptors for all mapping images.
Additionally, \cite{sarlin21pixloc} uses DenseVLAD \cite{torii2015denseVLAD} to retrieve the top reference images for each query image, together with the scene point cloud to associate 3D coordinates to the pixels of the reference images.
Thus, in Table 1 and Table 2 of the main paper we approximate the map size of pixLoc with the average memory demand of storing all mapping images, as the additional space required to store retrieval descriptors and point cloud is tiny in comparison.

\subsubsection{hybridSC \cite{compression2019cvpr}}

\paragraph{Mapping Time.} hybridSC is a compression algorithm that reduces the average size of feature descriptors.
It constitutes a post-processing on top of existing 3D map reconstructions.
We ignore any overhead incurred by descriptor compression, and report the average triangulation time of COLMAP \cite{schoenberger2016sfm} using SIFT \cite{sift}, see our reasoning for Active Search above.
This is the number we give in Table 2.
 
\paragraph{Map Size.}
We report the average memory demand per scene given in \cite{compression2019cvpr} in Table 2 of the main paper.

\subsubsection{GoMatch \cite{zhou2022gomatch}}

\paragraph{Mapping Time.}
GoMatch estimates a pose by matching constellations of 2D feature detections against a 3D point cloud.
Thus, as mapping time in Table 2 of the main paper, we report the average triangulation time of COLMAP \cite{schoenberger2016sfm} using SIFT \cite{sift} features, see our reasoning for Active Search above.

\paragraph{Map Size.}
We report the average memory demand per scene given in \cite{zhou2022gomatch} in Table 2 of the main paper.

\subsubsection{PoseNet17 \cite{Kendall2017GeometricLF}}

\paragraph{Mapping Time.}
As mapping time in Table 2 of the main paper, we report the range of training times given in \cite{Kendall2017GeometricLF}.

\paragraph{Map Size.}
We report the memory demand given in \cite{compression2019cvpr}.

\subsubsection{MS-Transformer \cite{Shavit2021MStransformer}}

\paragraph{Mapping Time.}
We used the code released by the authors of the paper \cite{Shavit2021MStransformer} to get an estimate of the time required to train a Multi-Scene Transformer network on the four scenes from the Cambridge Landmarks dataset~\cite{kendall2015posenet} they report results for (the paper doesn't report results for the ``Great Court'' scene).
We found that, on a machine equipped with a Nvidia RTX A6000 GPU, training the model for 1 epoch takes approximately 3 minutes.
From the paper we saw that, on the Cambridge dataset, the training should run for 550 epochs initially, followed by 40 epochs in which only the position branch is optimized.
This means that the network would be fully trained in approximately 29.5h but, as the same network can then be used to localize in all four scenes of the dataset, in Table~2 of the main text we report the ``average'' per-scene time of 7h.

\paragraph{Map Size.}
As per the paper~\cite{Shavit2021MStransformer}, the trained network has a memory footprint of 74.6MB.
Since the network can be used to localize in all four scenes of the Cambridge Landmarks dataset considered by the authors, in the main text we report an ``average'' per-scene map size of 18MB.

\subsubsection{SANet \cite{yang2019sanet}}

\paragraph{Mapping Time.}
The SANet paper uses a scene-agnostic network to interpolate the coordinate maps associated to the top-scoring images returned by a retrieval method such as NetVLAD~\cite{netvlad}.
As such we can consider the time it takes to create the retrieval index for the set training images as the mapping time.
In \cite{yang2019sanet}, the authors report that each forward pass of NetVLAD takes approximately 0.06s.
We can compute the average mapping time per scene by simply multiplying the above by the number of frames in the training sets of the various datasets.
For the indoor datasets (7 Scenes\cite{shotton2013scene} and 12 Scenes\cite{valentin2016learning}) the average scan length is 2259 frames.
As such, we report in Table~1 of the main paper a mapping time of $\sim$2.3min.
For the Cambridge Landmarks dataset\cite{kendall2015posenet} the average training scan length is 1073 frames, so we report a mapping time of $\sim$1min in Table~2.

\paragraph{Map Size.}
According to \cite{yang2019sanet}, for each element of the training set, SANet requires to store the (resized) RGB-D image, its pose, and the corresponding NetVLAD description, for a total of 248kB.
Similarly to the previous paragraph, we use the avg.~length of the training scans in both indoor and outdoor datasets to derive the ``average'' size of SANet maps.
In the main paper we thus report $\sim$550MB for the indoor maps and $\sim$260MB for the outdoor maps.

\subsubsection{SRC \cite{dong2022visual}}

\paragraph{Mapping Time.}
The Scene Region Classification approach\cite{dong2022visual} is based on a neural network classifying each pixel of the query images into one of the discrete regions determined via hierarchical partitioning of RGB-D mapping images.
As such, the mapping time is comprised of two components: (a) the time required to create the region partitions, and (b) the training of the Scene Region Classification network, learning to classify pixels in the input images into their respective region.
In \cite{dong2022visual} the authors claim that the training of the region classification network converges in $\sim$2min, so we report that in Tables 1 and 2 of the main text, but that doesn't include the time required to partition the training data.
In our experiments with the code publicly released by the authors we saw that hierarchical partitioning takes between 1 and 2 minutes per scene (on 7 Scenes~\cite{kendall2015posenet}), and that ought to be added to the mapping time.

\paragraph{Map Size.}
In \cite{dong2022visual} the authors report that a trained network occupies $\sim$40MB.

\subsection{Wayspots Dataset}
\label{supp:wayspots}

\subsubsection{Scene Selection}

We take the 10 scenes of our Wayspots dataset from the training split of the MapFree \cite{arnold2022mapfree} dataset.
The MapFree dataset is designed for relative pose regression from pairs of images.
As training data, Arnold \etal \cite{arnold2022mapfree} provide 460 outdoor scenes with two fully registered scans each.
These scans are meant as source for sampling image pairs to train relative pose regression.
We re-purpose this data for standard visual relocalisation using one scan as mapping sequence, and using the second scan for queries.
We reserve the first 200 scenes of the MapFree training corpus for training an ACE backbone variant, see Sec.~\ref{supp:exp:backbone}.
For our Wayspots dataset, we selected the scenes \emph{s00200} - \emph{s00209}. 

\subsubsection{Pseudo Ground Truth}

The MapFree dataset consists of scans from phones that ran RGB-based visual odometry to track camera poses.
Arnold \etal \cite{arnold2022mapfree} triangulated each scan using the phone's camera poses, and registered scans of the same scene using COLMAP \cite{schoenberger2016sfm}.
They also refined camera poses using bundle adjustment.
The bundle-adjusted SfM poses are what the MapFree dataset provides as pseudo ground truth.

However, we are interested in using the original phone poses for mapping since they can be generated in real-time, and thus incur practically no further mapping overhead.
Arnold \etal \cite{arnold2022mapfree} kindly provided the original phone poses based on our request.
We can use the phone poses of the mapping scan for creating a scene representation.
However, phone poses of the query scan are not registered to the mapping poses.
To enable evaluation, we register the phone poses of the mapping scan to their SfM counterpart poses, essentially registering them to the SfM query poses. 
In doing so, we regard the SfM poses as the unobservable pseudo ground truth, and the phone poses as the less precise -- but observable -- mapping input.

Note: Although our dataset does not use SfM poses for mapping, SfM-based relocalizers might still have an advantage on these scenes.
The MapFree dataset only contains scenes where COLMAP triangulation and registration was successful.
Thus, there might be a selection bias in favour of sparse feature-based methods.
In this work, we only compare scene coordinate regression approaches on this dataset where we do not expect a bias towards one method \cite{brachmann2021limits}.

\subsubsection{Phone-to-SfM Pose Alignment}

\begin{figure}[t]
  \centering
   \includegraphics[width=1.0\linewidth]{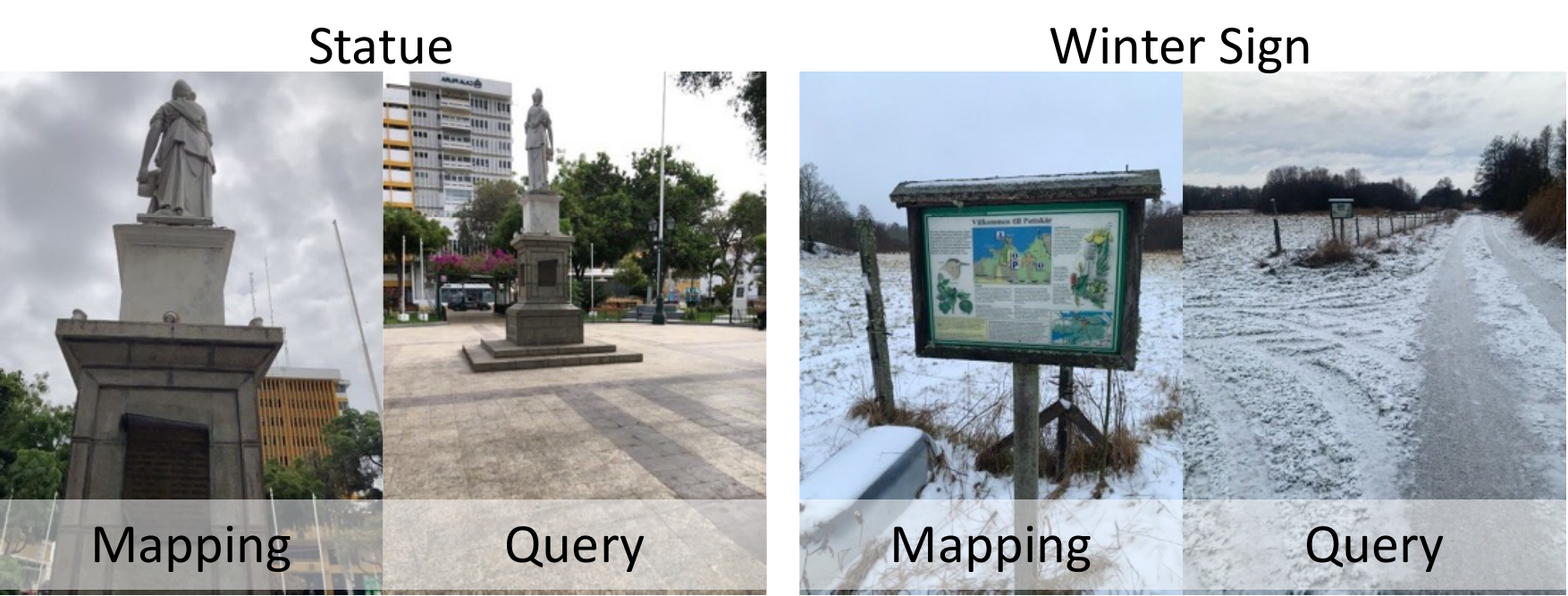}
   \caption{\textbf{Hard Scenes.} Example frames from the two hardest scenes of the Wayspots dataset. These scenes where mapped from close-up but queried from afar.}
   \vspace{-0.5cm}
   \label{fig:supp:hard_frames}
\end{figure}

\begin{table*}[]
\centering
\begin{tabular}{@{}lcccccccc@{}}
\toprule
 &
  \multicolumn{2}{c}{Number of Frames} &
   &
  \multicolumn{2}{c}{Phone-to-SfM Alignment Residuals} &
   &
  \multicolumn{2}{c}{ACE Results (10cm,5$^\circ$)} \\ \cmidrule(lr){2-3} \cmidrule(lr){5-6} \cmidrule(l){8-9} 
Scene &
  Mapping &
  Query &
   &
  \begin{tabular}[c]{@{}c@{}}Position \\ (mean/median/max)\end{tabular} &
  \begin{tabular}[c]{@{}c@{}}Rotation\\ (mean/median/max)\end{tabular} &
   &
  \begin{tabular}[c]{@{}c@{}}Mapping  w/\\ Phone Poses\end{tabular} &
  \begin{tabular}[c]{@{}c@{}}Mapping w/\\ SfM Poses\end{tabular} \\ \midrule
Cubes        & 581 & 575 &  & 2.6cm/1.9cm/8.6cm   & 0.6$^\circ$/0.6$^\circ$/0.9$^\circ$ &  & 97.0\% & 97.6\% \\
Bears        & 581 & 580 &  & 1.3cm/1.2cm/3.1cm   & 0.5$^\circ$/0.5$^\circ$/0.9$^\circ$ &  & 80.7\% & 80.9\% \\
Winter Sign  & 580 & 580 &  & 2.1cm/1.4cm/13.8cm  & 0.6$^\circ$/0.6$^\circ$/2.2$^\circ$ &  & 1.0\%  & 0.7\%  \\
Inscription  & 578 & 555 &  & 4.4cm/3.7cm/21.5cm  & 1.0$^\circ$/1.0$^\circ$/1.5$^\circ$ &  & 49.0\% & 59.5\% \\
The Rock     & 579 & 578 &  & 5.1cm/5.5cm/10.2cm  & 0.7$^\circ$/0.8$^\circ$/1.1$^\circ$ &  & 100\%  & 100\%  \\
Tendrils     & 580 & 581 &  & 8.4cm/4.6cm/26.9cm  & 1.7$^\circ$/1.7$^\circ$/2.0$^\circ$ &  & 34.9\% & 34.6\% \\
Map          & 575 & 503 &  & 7.2cm/4.5cm/23.2cm  & 1.0$^\circ$/0.9$^\circ$/1.9$^\circ$ &  & 56.5\% & 87.7\% \\
Square Bench & 580 & 577 &  & 8.4cm/5.9cm/22.1cm  & 0.7$^\circ$/0.6$^\circ$/1.0$^\circ$ &  & 66.7\% & 93.9\% \\
Statue       & 560 & 553 &  & 14.5cm/9.3cm/43.6cm & 0.5$^\circ$/0.5$^\circ$/1.2$^\circ$ &  & 0\%    & 0\%    \\
Lawn         & 575 & 579 &  & 3.3cm/3.2cm/6.2cm   & 0.4$^\circ$/0.5$^\circ$/0.9$^\circ$ &  & 35.8\% & 40.4\% \\ \midrule
Average      &     &     &  &                     &             &  & 52.2\% & 59.5\% \\ \bottomrule
\end{tabular}
\caption{\textbf{Wayspots Pose Alignment.} Evaluation of the alignment quality between the phone's poses (estimated using visual odometry, in real time) and bundle-adjusted SfM poses obtained offline (used as pseudo ground truth to evaluate camera relocalization results) on the Wayspots dataset.
We also show, on the right, the difference in performance achieved by ACE when using phone poses, vs.~using SfM poses during mapping.
In Figure 4 of the main paper we report the results obtained using the former, as that's ACE's intended use case.
}
\label{tab:mapfree}
\end{table*}

To register phone poses to SfM poses, we use the Kabsch algorithm \cite{kabsch1976solution} within a RANSAC loop \cite{fischler1981random}, relying on 3D camera positions alone.
We sample 1000 random triplets of corresponding camera positions to generate registration hypotheses.
We rank hypotheses by inlier counting with a threshold of 10cm.
We re-fit the winning hypotheses to all inlier correspondences.
Note that Arnold \etal have already re-scaled the SfM poses according to the phone trajectories after bundle adjustment to prevent scale drift \cite{arnold2022mapfree}.

Due to SfM poses being bundle-adjusted, we cannot expect perfect alignment with the original phone poses.
We list alignment errors for each scene in Table \ref{tab:mapfree}.
Since we do not use camera rotation in our pose registration, statistics over rotation alignment provide a good plausibility check on the success of our approach.
We see that some scenes exhibit noticeable drift between poses from phone tracking and poses from SfM.
However, the median alignment error is below 10cm for all scenes, such that at least 50\% relocalisation rate is achievable for all scenes using our proposed error threshold of 10cm and 5$^\circ$.
Better results are possible if a method refines mapping poses during mapping. 

To verify that alignment errors are not the main cause of poor results on some scenes, we repeat experiments with ACE using SfM poses for mapping, see Table \ref{tab:mapfree}.
While results do indeed improve, they improve only moderately on average.
Difficult scenes remain difficult (such as ``Winter Sign'' and ``Statue'') even when factoring out alignment errors of the pseudo ground truth.
See Fig.~\ref{fig:supp:hard_frames} for example frames of these two scenes, highlighting their challenges.

\subsection{Gradient Decorrelation}
\label{supp:exp:decorrelation}

\begin{table*}[]
\centering
\footnotesize
\begin{tabular}{@{}lcclcclcclcc@{}}
\toprule
      &              &                              &  & \multicolumn{2}{c}{7Scenes} &  & \multicolumn{2}{c}{12Scenes} &  & Cambridge  &        \\ \cmidrule(lr){5-6} \cmidrule(lr){8-9}
\multirow{-2}{*}{Method} &
  \multirow{-2}{*}{Gradients} &
  \multirow{-2}{*}{\begin{tabular}[c]{@{}c@{}}Mapping \\ Time\end{tabular}} &
   &
  SfM poses &
  D-SLAM poses &
   &
  SfM poses &
  D-SLAM poses &
   &
  (avg. median error) &
  \multirow{-2}{*}{Wayspots} \\ \midrule
DSAC* & correlated   & \redcell 15h  &  & 96.0\%       & 81.1\%       &  & 99.6\%        & 98.8\%       &  & 19cm/0.4°  & 50.7\% \\
DSAC* & correlated   & \greencell 5min &  & 3.6\%        & 5.5\%        &  & 0.0\%         & 0.2\%        &  & 2375cm/49° & 7.4\%  \\
\rowcolor[HTML]{E0E0E0} 
ACE   & correlated   & \greencell 5min &  & 91.2\%       & 72.9\%       &  & 74.3\%        & 75.2\%       &  & 406cm/8.7° & 49.7\% \\
ACE   & decorrelated & \greencell 5min &  & 97.1\%       & 80.8\%       &  & 99.9\%        & 99.6\%       &  & 25cm/0.4°  & 52.2\% \\ \bottomrule
\end{tabular}
\caption{\textbf{Gradient Decorrelation.} The gray line marks a version of ACE where the training buffer is shuffled image-wise instead of feature-wise, akin to the training procedure of DSAC*. Gradient decorrelation improves accuracy and stability of ACE.}
\label{supp:tab:gradients}
\end{table*}

We train ACE with correlated gradients to show the detrimental effect on accuracy and stability. 
Akin to DSAC* \cite{Brachmann2021dsacstar}, we train ACE by using all features per mapping image, and by shuffling the ACE buffer image-wise.
In this variant, ACE sees correlated features from a single mapping image during each training iteration, as opposed to random features selected from the entire mapping sequence. 
Accuracy deteriorates across all datasets, see Table ~\ref{supp:tab:gradients}.
The effect is particularly pronounced on Cambridge where individual scenes, such as St.~Marys Church, diverge completely. 
Note that ACE with correlated gradients still exceeds the 5 minute version of DSAC*.
ACE starts with strong backbone features, and does more training iterations in the same time due to the small network.

\subsection{Feature Backbone}
\label{supp:exp:backbone}

We combine ACE training with some popular feature backbones and report results in Table~\ref{tab:backbone}.
Training a backbone specifically for the task of scene coordinate regression achieves best results.
Descriptor backbones of sparse feature-matching pipelines use 128 or 256 feature dimensions for faster matching, and leaner storage.
In contrast, scene coordinate regression can benefit from higher dimensional features with neglectable impact on computation time or memory.
Training our backbone on ScanNet\cite{dai2017scannet} yields best results.
We also trained a backbone variant on the MapFree dataset \cite{arnold2022mapfree}.
We used scenes \emph{s00000} to \emph{s00099}\footnote{We reserved the first 200 scenes of the MapFree training set for encoder experiments but ended up only using the first 100 scenes.}, and both available scans per scene since they are registered.
The MapFree backbone performs slightly worse than the ScanNet backbone.
Presumably ScanNet scenes, being indoors, contain less uninformative areas like sky or ground.

\begin{table}[]
\centering
\begin{tabular}{@{}lcc@{}}
\toprule
Backbone                             & \begin{tabular}[c]{@{}c@{}}Descriptor \\ Dim.\end{tabular} & \begin{tabular}[c]{@{}c@{}}Accuracy \\ (7Scenes, 5cm5$^\circ$)\end{tabular} \\ \midrule
DenseSIFT \cite{liu2011siftflow}                & 128                                                        & 83.0\%                                                                 \\
DISK \cite{Tyszkiewicz2020DISK}      & 128                                                        & 73.6\%                                                                 \\
R2D2 \cite{revaud2019r2d2}           & 128                                                        & 87.7\%                                                                 \\
SuperPoint \cite{detone18superpoint} & 256                                                        & 91.6\%                                                                 \\ \midrule
ACE {[}MapFree \cite{arnold2022mapfree}{]}                    & 512                                                        & 91.5\%                                                                 \\ \midrule
\multirow{3}{*}{ACE {[}ScanNet \cite{dai2017scannet}{]}}   & 128                                                        & 95.8\%                                                                   \\
                                     & 256                                                        & 96.4\%                                                                   \\
                                     & 512                                                        & \textbf{97.1}\%                                                                 \\ \bottomrule
\end{tabular}
\caption{\textbf{Backbones.} We can use multiple off-the-shelf backbones with ACE. Our backbone trained on ScanNet achieves best results.}
\label{tab:backbone}
\end{table}

\subsection{Map Visualization}
\label{supp:map_vis}

To create 3D representations of the maps learned by ACE (as shown in Fig.~4 and 5 of the main paper), we run the mapping sequence through the learned network and accumulate 3D scene coordinate predictions. 
To obtain colored point clouds, we rescale the mapping images to the network output resolution, and read out the color of each predicted point.
To de-clutter the point clouds, we remove coordinate predictions that are more than 10m from the image plane.
To visualize camera trajectories, we connect consecutive camera positions with a line.
We skip lines when cameras are further than 0.5m apart, assuming an outlier estimate.
To also show camera orientation, we place a camera frustum for 1 out of 25 frames.

{\small
\bibliographystyle{ieee_fullname}
\bibliography{egbib}

\begin{thebibliography}{10}\itemsep=-1pt

\bibitem{arkit}
Apple.
\newblock
  \href{https://developer.apple.com/documentation/arkit/configuration_objects/understanding_world_tracking}{ARKit}.
\newblock {Accessed: 11 November 2022}.

\bibitem{netvlad}
Relja Arandjelovic, Petr Gronat, Akihiko Torii, Tomas Pajdla, and Josef Sivic.
\newblock {NetVLAD: CNN} architecture for weakly supervised place recognition.
\newblock In {\em CVPR}, 2016.

\bibitem{arnold2022mapfree}
Eduardo Arnold, Jamie Wynn, Sara Vicente, Guillermo Garcia-Hernando, {\'{A}}ron
  Monszpart, Victor~Adrian Prisacariu, Daniyar Turmukhambetov, and Eric
  Brachmann.
\newblock Map-free visual relocalization: Metric pose relative to a single
  image.
\newblock In {\em ECCV}, 2022.

\bibitem{kmeans++}
David Arthur and Sergei Vassilvitskii.
\newblock K-means++: The advantages of careful seeding.
\newblock volume~8, pages 1027--1035, 01 2007.

\bibitem{brachmann2021limits}
Eric Brachmann, Martin Humenberger, Carsten Rother, and Torsten Sattler.
\newblock On the limits of pseudo ground truth in visual camera
  re-localisation.
\newblock In {\em ICCV}, 2021.

\bibitem{brachmann2017dsac}
Eric Brachmann, Alexander Krull, Sebastian Nowozin, Jamie Shotton, Frank
  Michel, Stefan Gumhold, and Carsten Rother.
\newblock {DSAC}-differentiable ransac for camera localization.
\newblock In {\em CVPR}, 2017.

\bibitem{brachmann2016}
Eric Brachmann, Frank Michel, Alexander Krull, Michael~Y. Yang, Stefan Gumhold,
  and Carsten Rother.
\newblock Uncertainty-driven 6{D} pose estimation of objects and scenes from a
  single {RGB} image.
\newblock In {\em CVPR}, 2016.

\bibitem{Brachmann2018dsacpp}
Eric Brachmann and Carsten Rother.
\newblock {Learning Less is More - 6D Camera Localization via 3D Surface
  Regression}.
\newblock In {\em CVPR}, 2018.

\bibitem{Brachmann2019ESAC}
Eric Brachmann and Carsten Rother.
\newblock Expert sample consensus applied to camera re-localization.
\newblock In {\em ICCV}, 2019.

\bibitem{brachmann2019NGransac}
Eric Brachmann and Carsten Rother.
\newblock Neural-guided {RANSAC}: Learning where to sample model hypotheses.
\newblock In {\em ICCV}, 2019.

\bibitem{Brachmann2021dsacstar}
Eric Brachmann and Carsten Rother.
\newblock Visual camera re-localization from {RGB} and {RGB-D} images using
  {DSAC}.
\newblock {\em TPAMI}, 2021.

\bibitem{Brahmbhatt2018mapnet}
Samarth Brahmbhatt, Jinwei Gu, Kihwan Kim, James Hays, and Jan Kautz.
\newblock Geometry-aware learning of maps for camera localization.
\newblock In {\em CVPR}, 2018.

\bibitem{compression2019cvpr}
Federico Camposeco, Andrea Cohen, Marc Pollefeys, and Torsten Sattler.
\newblock Hybrid scene compression for visual localization.
\newblock In {\em CVPR}, 2019.

\bibitem{Cavallari2019network}
Tommaso Cavallari, Luca Bertinetto, Jishnu Mukhoti, Philip Torr, and Stuart
  Golodetz.
\newblock Let's take this online: Adapting scene coordinate regression network
  predictions for online rgb-d camera relocalisation.
\newblock In {\em 3DV}, 2019.

\bibitem{cavallari2017fly}
Tommaso Cavallari, Stuart Golodetz, Nicholas~A Lord, Julien Valentin, Luigi
  Di~Stefano, and Philip~HS Torr.
\newblock On-the-fly adaptation of regression forests for online camera
  relocalisation.
\newblock In {\em CVPR}, 2017.

\bibitem{Cavallari2019cascade}
Tommaso {Cavallari}, Stuart {Golodetz}, Nicholas~A. {Lord}, Julien {Valentin},
  Victor~A. {Prisacariu}, Luigi {Di Stefano}, and Philip H.~S. {Torr}.
\newblock Real-time rgb-d camera pose estimation in novel scenes using a
  relocalisation cascade.
\newblock {\em TPAMI}, 2019.

\bibitem{Chelani2021privacyhack}
Kunal Chelani, Fredrik Kahl, and Torsten Sattler.
\newblock How privacy-preserving are line clouds? recovering scene details from
  3d lines.
\newblock In {\em CVPR}, 2021.

\bibitem{dai2017scannet}
Angela Dai, Angel~X Chang, Manolis Savva, Maciej Halber, Thomas Funkhouser, and
  Matthias Nie{\ss}ner.
\newblock {ScanNet}: Richly-annotated {3D} reconstructions of indoor scenes.
\newblock In {\em CVPR}, 2017.

\bibitem{dai2017bundlefusion}
Angela Dai, Matthias Nie{\ss}ner, Michael Zollh{\"o}fer, Shahram Izadi, and
  Christian Theobalt.
\newblock Bundlefusion: Real-time globally consistent {3D} reconstruction using
  on-the-fly surface reintegration.
\newblock {\em TOG}, 2017.

\bibitem{detone18superpoint}
Daniel DeTone, Tomasz Malisiewicz, and Andrew Rabinovich.
\newblock Superpoint: Self-supervised interest point detection and description.
\newblock In {\em CVPRW}, 2018.

\bibitem{dong2022visual}
Siyan Dong, Shuzhe Wang, Yixin Zhuang, Juho Kannala, Marc Pollefeys, and
  Baoquan Chen.
\newblock Visual localization via few-shot scene region classification.
\newblock In {\em 3DV}, 2022.

\bibitem{fischler1981random}
Martin~A Fischler and Robert~C Bolles.
\newblock Random sample consensus: a paradigm for model fitting with
  applications to image analysis and automated cartography.
\newblock {\em Communications of the ACM}, 1981.

\bibitem{gao2003complete}
Xiao-Shan Gao, Xiao-Rong Hou, Jianliang Tang, and Hang-Fei Cheng.
\newblock Complete solution classification for the perspective-three-point
  problem.
\newblock {\em TPAMI}, 2003.

\bibitem{gcpgpupricing}
Google.
\newblock \href{https://cloud.google.com/compute/gpus-pricing}{Google Compute
  Engine GPU Pricing}.
\newblock {Accessed: 11 November 2022}.

\bibitem{arcore}
Google.
\newblock \href{https://developers.google.com/ar/develop/fundamentals}{ARCore}.
\newblock {Accessed: 11 November 2022}.

\bibitem{kapture2020}
Martin Humenberger, Yohann Cabon, Nicolas Guerin, Julien Morat, Jérôme
  Revaud, Philippe Rerole, Noé Pion, Cesar de Souza, Vincent Leroy, and
  Gabriela Csurka.
\newblock Robust image retrieval-based visual localization using {Kapture},
  2020.

\bibitem{Irschara2009sfmfast}
Arnold Irschara, Christopher Zach, Jan-Michael Frahm, and Horst Bischof.
\newblock From structure-from-motion point clouds to fast location recognition.
\newblock In {\em CVPR}, 2009.

\bibitem{izadi2011kinectfusion}
Shahram Izadi, David Kim, Otmar Hilliges, David Molyneaux, Richard Newcombe,
  Pushmeet Kohli, Jamie Shotton, Steve Hodges, Dustin Freeman, Andrew Davison,
  et~al.
\newblock {Kinectfusion: real-time 3D reconstruction and interaction using a
  moving depth camera}.
\newblock In {\em UIST}, 2011.

\bibitem{kabsch1976solution}
Wolfgang Kabsch.
\newblock A solution for the best rotation to relate two sets of vectors.
\newblock {\em Acta Crystallographica Section A: Crystal Physics, Diffraction,
  Theoretical and General Crystallography}, 1976.

\bibitem{Kendall2017GeometricLF}
Alex Kendall and Roberto Cipolla.
\newblock Geometric loss functions for camera pose regression with deep
  learning.
\newblock {\em CVPR}, 2017.

\bibitem{kendall2015posenet}
Alex Kendall, Matthew Grimes, and Roberto Cipolla.
\newblock Posenet: A convolutional network for real-time 6-{DOF} camera
  relocalization.
\newblock In {\em CVPR}, 2015.

\bibitem{Levenberg44lm}
K. Levenberg.
\newblock A method for the solution of certain problems in least squares.
\newblock {\em Quaterly Journal on Applied Mathematics}, 1944.

\bibitem{li2020hierarchical}
Xiaotian Li, Shuzhe Wang, Yi Zhao, Jakob Verbeek, and Juho Kannala.
\newblock Hierarchical scene coordinate classification and regression for
  visual localization.
\newblock In {\em CVPR}, 2020.

\bibitem{liu2011siftflow}
Ce Liu, Jenny Yuen, and Antonio Torralba.
\newblock {SIFT} flow: Dense correspondence across scenes and its applications.
\newblock {\em TPAMI}, 2011.

\bibitem{fcn2015}
J. Long, E. Shelhamer, and T. Darrell.
\newblock Fully convolutional networks for semantic segmentation.
\newblock In {\em CVPR}, 2015.

\bibitem{Loshchilov2019DecoupledWD}
Ilya Loshchilov and Frank Hutter.
\newblock Decoupled weight decay regularization.
\newblock In {\em ICLR}, 2019.

\bibitem{sift}
David~G. Lowe.
\newblock Distinctive image features from scale-invariant keypoints.
\newblock {\em IJCV}, 2004.

\bibitem{lynen2020visin}
Simon Lynen, Bernhard Zeisl, Dror Aiger, Michael Bosse, Joel Hesch, Marc
  Pollefeys, Roland Siegwart, and Torsten Sattler.
\newblock Large-scale, real-time visual–inertial localization revisited.
\newblock {\em Intl. Journal of Robotics Research}, 2020.

\bibitem{maggio2022locnerf}
Dominic Maggio, Marcus Abate, Jingnan Shi, Courtney Mario, and Luca Carlone.
\newblock Loc-{NeRF}: Monte carlo localization using neural radiance fields,
  2022.

\bibitem{Marquardt63lm}
Donald~W. Marquardt.
\newblock An algorithm for least-squares estimation of nonlinear parameters.
\newblock {\em Journal of the Society for Industrial and Applied Mathematics},
  1963.

\bibitem{mildenhall2020nerf}
Ben Mildenhall, Pratul~P. Srinivasan, Matthew Tancik, Jonathan~T. Barron, Ravi
  Ramamoorthi, and Ren Ng.
\newblock Nerf: Representing scenes as neural radiance fields for view
  synthesis.
\newblock In {\em ECCV}, 2020.

\bibitem{newcombe2011kinectfusion}
Richard~A Newcombe, Shahram Izadi, Otmar Hilliges, David Molyneaux, David Kim,
  Andrew~J Davison, Pushmeet Kohi, Jamie Shotton, Steve Hodges, and Andrew
  Fitzgibbon.
\newblock {KinectFusion}: Real-time dense surface mapping and tracking.
\newblock In {\em ISMAR}, 2011.

\bibitem{Panek2022meshloc}
Vojtech Panek, Zuzana Kukelova, and Torsten Sattler.
\newblock {MeshLoc: Mesh-Based Visual Localization}.
\newblock In {\em ECCV}, 2022.

\bibitem{paszke2017automatic}
Adam Paszke, Sam Gross, Soumith Chintala, Gregory Chanan, Edward Yang, Zachary
  DeVito, Zeming Lin, Alban Desmaison, Luca Antiga, and Adam Lerer.
\newblock Automatic differentiation in {PyTorch}.
\newblock In {\em NIPS-W}, 2017.

\bibitem{revaud2019apgem}
Jerome Revaud, Jon Almazan, Rafael Rezende, and Cesar~De Souza.
\newblock Learning with average precision: Training image retrieval with a
  listwise loss.
\newblock In {\em ICCV}, 2019.

\bibitem{revaud2019r2d2}
Jerome Revaud, Philippe Weinzaepfel, C{\'{e}}sar~Roberto de Souza, and Martin
  Humenberger.
\newblock {R2D2:} repeatable and reliable detector and descriptor.
\newblock In {\em NeurIPS}, 2019.

\bibitem{sarlin2019coarse}
Paul-Edouard Sarlin, Cesar Cadena, Roland Siegwart, and Marcin Dymczyk.
\newblock From coarse to fine: Robust hierarchical localization at large scale.
\newblock In {\em CVPR}, 2019.

\bibitem{sarlin2020superglue}
Paul-Edouard Sarlin, Daniel DeTone, Tomasz Malisiewicz, and Andrew Rabinovich.
\newblock {SuperGlue}: Learning feature matching with graph neural networks.
\newblock In {\em CVPR}, 2020.

\bibitem{sarlin21pixloc}
Paul-Edouard Sarlin, Ajaykumar Unagar, Måns Larsson, Hugo Germain, Carl Toft,
  Victor Larsson, Marc Pollefeys, Vincent Lepetit, Lars Hammarstrand, Fredrik
  Kahl, and Torsten Sattler.
\newblock {Back to the Feature: Learning Robust Camera Localization from Pixels
  to Pose}.
\newblock In {\em CVPR}, 2021.

\bibitem{sattler2011fast}
Torsten Sattler, Bastian Leibe, and Leif Kobbelt.
\newblock {Fast image-based localization using direct 2D-to-3D matching}.
\newblock In {\em ICCV}, 2011.

\bibitem{sattler2012improving}
Torsten Sattler, Bastian Leibe, and Leif Kobbelt.
\newblock Improving image-based localization by active correspondence search.
\newblock In {\em ECCV}, 2012.

\bibitem{Sattler2017AS}
Torsten Sattler, Bastian Leibe, and Leif Kobbelt.
\newblock {Efficient \& Effective Prioritized Matching for Large-Scale
  Image-Based Localization}.
\newblock {\em TPAMI}, 2017.

\bibitem{sattler2019limits}
Torsten Sattler, Qunjie Zhou, Marc Pollefeys, and Laura Leal-Taixe.
\newblock Understanding the limitations of cnn-based absolute camera pose
  regression.
\newblock In {\em CVPR}, 2019.

\bibitem{schoenberger2016sfm}
Johannes~Lutz Sch\"{o}nberger and Jan-Michael Frahm.
\newblock Structure-from-motion revisited.
\newblock In {\em CVPR}, 2016.

\bibitem{Shavit2021MStransformer}
Yoli Shavit, Ron Ferens, and Yosi Keller.
\newblock Learning multi-scene absolute pose regression with transformers.
\newblock In {\em ICCV}, 2021.

\bibitem{shotton2013scene}
Jamie Shotton, Ben Glocker, Christopher Zach, Shahram Izadi, Antonio Criminisi,
  and Andrew Fitzgibbon.
\newblock Scene coordinate regression forests for camera relocalization in
  {RGB-D} images.
\newblock In {\em CVPR}, 2013.

\bibitem{Smith2019cycliclr}
Leslie~N. Smith and Nicholay Topin.
\newblock {Super-convergence: very fast training of neural networks using large
  learning rates}.
\newblock In {\em Artificial Intelligence and Machine Learning for Multi-Domain
  Operations Applications}, 2019.

\bibitem{Snavely2006bundler}
Noah Snavely, Steven~M. Seitz, and Richard Szeliski.
\newblock Photo tourism: Exploring photo collections in 3d.
\newblock In {\em SIGGRAPH}, 2006.

\bibitem{Speciale2019privacy}
Pablo Speciale, Johannes~L. Schonberger, Sing~Bing Kang, Sudipta~N. Sinha, and
  Marc Pollefeys.
\newblock Privacy preserving image-based localization.
\newblock In {\em CVPR}, June 2019.

\bibitem{torii2015denseVLAD}
Akihiko Torii, Relja Arandjelovic, Josef Sivic, Masatoshi Okutomi, and Tomas
  Pajdla.
\newblock 24/7 place recognition by view synthesis.
\newblock In {\em CVPR}, 2015.

\bibitem{turkoglu2021visual}
Mehmet~{\"{O}}zg{\"{u}}r T{\"{u}}rko\u{g}lu, Eric Brachmann, Konrad Schindler,
  Gabriel Brostow, and \'{A}ron Monszpart.
\newblock {Visual Camera Re-Localization Using Graph Neural Networks and
  Relative Pose Supervision}.
\newblock In {\em 3DV}, 2021.

\bibitem{Tyszkiewicz2020DISK}
Micha\l Tyszkiewicz, Pascal Fua, and Eduard Trulls.
\newblock {DISK}: Learning local features with policy gradient.
\newblock In {\em NeurIPS}, 2020.

\bibitem{valentin2016learning}
Julien Valentin, Angela Dai, Matthias Nie{\ss}ner, Pushmeet Kohli, Philip Torr,
  Shahram Izadi, and Cem Keskin.
\newblock Learning to navigate the energy landscape.
\newblock In {\em 3DV}, 2016.

\bibitem{valentin2015cvpr}
Julien Valentin, Matthias Nie{\ss}ner, Jamie Shotton, Andrew Fitzgibbon,
  Shahram Izadi, and Philip H.~S. Torr.
\newblock Exploiting uncertainty in regression forests for accurate camera
  relocalization.
\newblock In {\em CVPR}, 2015.

\bibitem{WinkelbauerICRA21}
Dominik Winkelbauer, Maximilian Denninger, and Rudolph Triebel.
\newblock Learning to localize in new environments from synthetic training
  data.
\newblock In {\em ICRA}, 2021.

\bibitem{wu2011visualsfm}
Changchang Wu.
\newblock {VisualSFM}: A visual structure from motion system, 2011.

\bibitem{yang2019sanet}
Luwei Yang, Ziqian Bai, Chengzhou Tang, Honghua Li, Yasutaka Furukawa, and Ping
  Tan.
\newblock {SANet}: {S}cene agnostic network for camera localization.
\newblock In {\em ICCV}, 2019.

\bibitem{Yang2022squeezer}
Luwei Yang, Rakesh Shrestha, Wenbo Li, Shuaicheng Liu, Guofeng Zhang, Zhaopeng
  Cui, and Ping Tan.
\newblock Scenesqueezer: Learning to compress scene for camera relocalization.
\newblock In {\em CVPR}, 2022.

\bibitem{yen2020inerf}
Lin Yen-Chen, Pete Florence, Jonathan~T. Barron, Alberto Rodriguez, Phillip
  Isola, and Tsung-Yi Lin.
\newblock {iNeRF}: Inverting neural radiance fields for pose estimation.
\newblock In {\em IROS}, 2021.

\bibitem{zhou2022gomatch}
Qunjie Zhou, S{\'e}rgio Agostinho, Aljo{\v{s}}a O{\v{s}}ep, and Laura
  Leal-Taix{\'e}.
\newblock Is geometry enough for matching in visual localization?
\newblock In {\em ECCV}, 2022.

\bibitem{zhou2020essnet}
Qunjie Zhou, Torsten Sattler, Marc Pollefeys, and Laura Leal-Taixe.
\newblock To learn or not to learn: Visual localization from essential
  matrices.
\newblock In {\em ICRA}, 2020.

\end{thebibliography}
}

\end{document}